%% file: acl_latex.tex
\pdfoutput=1

\documentclass[table,xcdraw,11pt]{article}

\usepackage[final]{acl}

\usepackage{times}
\usepackage{latexsym}

\usepackage[T1]{fontenc}

\usepackage[utf8]{inputenc}

\usepackage{microtype}

\usepackage{inconsolata}
\usepackage{booktabs}
\usepackage{graphicx}
\usepackage{subcaption}

\usepackage{multirow}

\usepackage{amssymb}
\usepackage{amsmath}
\usepackage{listings}
\usepackage[linesnumbered,ruled,vlined]{algorithm2e}

\usepackage{tcolorbox}

\title{Navi-\textit{plus}: Managing  Ambiguous GUI Navigation Tasks \\ with Follow-up Questions}

\author{Ziming Cheng\thanks{Equal contribution.}\thanks{Work done as an intern at SenseTime Research.} \\
Beijing University of Posts and Telecommunications
\And
Zhiyuan Huang\footnotemark[1] \\
SenseTime Research
\AND
Junting Pan \\
MMLab, CUHK
\And
Zhaohui Hou \\
SenseTime Research
\And
Mingjie Zhan \\
SenseTime Research \\}

\begin{document}
\maketitle
\begin{abstract}
Graphical user interfaces (GUI) automation agents are emerging as powerful tools, enabling humans to accomplish increasingly complex tasks on smart devices. However, users often inadvertently omit key information when conveying tasks, which hinders agent performance in the current agent paradigm that does not support immediate user intervention. To address this issue, we introduce a \textbf{Self-Supplement GUI Navigation} task that incorporates interactive information completion capabilities within GUI agents. We developed the \textbf{Navi-\textit{plus}} dataset with GUI follow-up question-answer pairs, alongside a \textbf{Dual-Stream Trajectory Evaluation} method to benchmark this new capability. Our results show that agents equipped with the ability to ask GUI follow-up questions can interact with human users and recover their performance when faced with ambiguous user tasks.
\end{abstract}

\input{latex/01-Introduction}

\input{latex/02-Related-Work}
\input{latex/figures/data-pipeline}
\input{latex/03-Self-Supplement-GUI-Navigation-Task}
\input{latex/04-Navi-plus-Dataset}

\input{latex/05-Evaluation-Methods}
\input{latex/06-Experimental-Setup}

\input{latex/tables/minus-2-datasets-small}
\input{latex/tables/ask-result-less-ac-m2w}
\input{latex/tables/across-model-scales-large}
\input{latex/07-Results-and-Discussion}

\input{latex/08-Conclusion}
\input{latex/09-Limitations}


\bibliography{custom}

\appendix

\newpage
\label{sec:appendix}

\input{latex/figures/data-distribution}

\input{latex/figures/hyperpara-androidcontrol-small}
\input{latex/figures/data-demo-androidcontrol}
\input{latex/figures/data-demo-mind2web}
\input{latex/Appendix/_Appendix-Overview}
\input{latex/Appendix/A-Qualitative-Examples-of-Naviplus-Dataset}
\input{latex/Appendix/B-Dataset-Statistics}
\input{latex/Appendix/C-Extended-Experiment-Results}
\input{latex/Appendix/D-Evaluation-Metrics-Formulas}

\input{latex/Appendix/E-Extanded-Related-Work}
\input{latex/Appendix/F-Prompt-Templates}
\input{latex/Appendix/G-Ethical-Considerations}

\end{document}

%% file: latex/01-Introduction.tex
\section{Introduction}
\label{intro}

Graphical User Interface (GUI) becomes the foundational approach of modern human-computer interaction with increasing numbers of screens filling people's lives. To augment human capabilities and mitigate mental burdens in operating digital devices, GUI automation agents have arisen in recent years. Following the advancements of (Multimodal) Large Language Models (LLMs or MLLMs), extensive efforts were invested in constructing these autonomous agents through large-scale continual pre-training (\citealp{cheng2024seeclick}, \citealp{chai2024amex}, \citealp{lin2024showui}), grounding-augmented supervised fine-tuning (\citealp{li2024androidcontrol}, \citealp{sun2024os-genesis}), and utilization of Chain-of-Action-Thought (CoAT) in navigation(\citealp{zhang2024AndroidintheZoo}, \citealp{liu2025infiguiagent}). 
\input{latex/figures/main}

\input{latex/figures/towards-proactive}

However, the previous paradigm of GUI navigation agents is limited to receiving full human instruction and performing actions serially, diminishing human control halfway through agent processing. Thus, a practical "elephant in the room" question arises: \textbf{\textit{If some important information is missing from the human instruction (i.e., product specs or important dates), how can the agent continue the task that it is expected to finish?}}

This issue has motivated us to review the formulation of the current GUI navigation task. In this paper, we propose a novel task called \textbf{Self-Supplement GUI Navigation}, which endows GUI agents with a new ability to handle the ambiguity of human instruction. The core idea is to add an "ASK" action in the agents' action space, enabling it to engage in intermediate natural language interactions with human users by proposing follow-up questions (Figure~\ref{fig:overview}).

We first design a data annotation pipeline to construct Navi\textit{plus} dataset from existing trajectory datasets using open-source LLMs(MLLMs). GUI navigation trajectories with ambiguous task descriptions are intentionally and controllably generated, along with corresponding GUI follow-up question-answering (QA) pairs.

We then include fair and comprehensive evaluation metrics to benchmark GUI agents' capability to complete the GUI navigation task when the task description is ambiguous. A \textbf{Dual-Stream Trajectory Evaluation} method is proposed to separately compute metrics for the operational actions (e.g., Step Success Rate) and for the additional ASK action, allowing direct model comparison.

Our experiments show that the missing information in task description can significantly harm GUI agents task success rate, but with the completion of ASK action, agents can restore up to 99.4\% of performance (Figure~\ref{fig:androidcontrol-compare}). Moreover, we find that modern MLLM-based GUI agents can seamlessly learn to propose follow-up questions and achieve satisfactory performance, with a timing accuracy of up to 0.947 and a content similarity of up to 0.832. Through extensive experiments, we further demonstrate that both model scale and dataset size affect performance on the proposed self-supplement GUI navigation task.

%% file: latex/figures/main.tex
\begin{figure}[t]
  \centering
  \includegraphics[width=\linewidth]{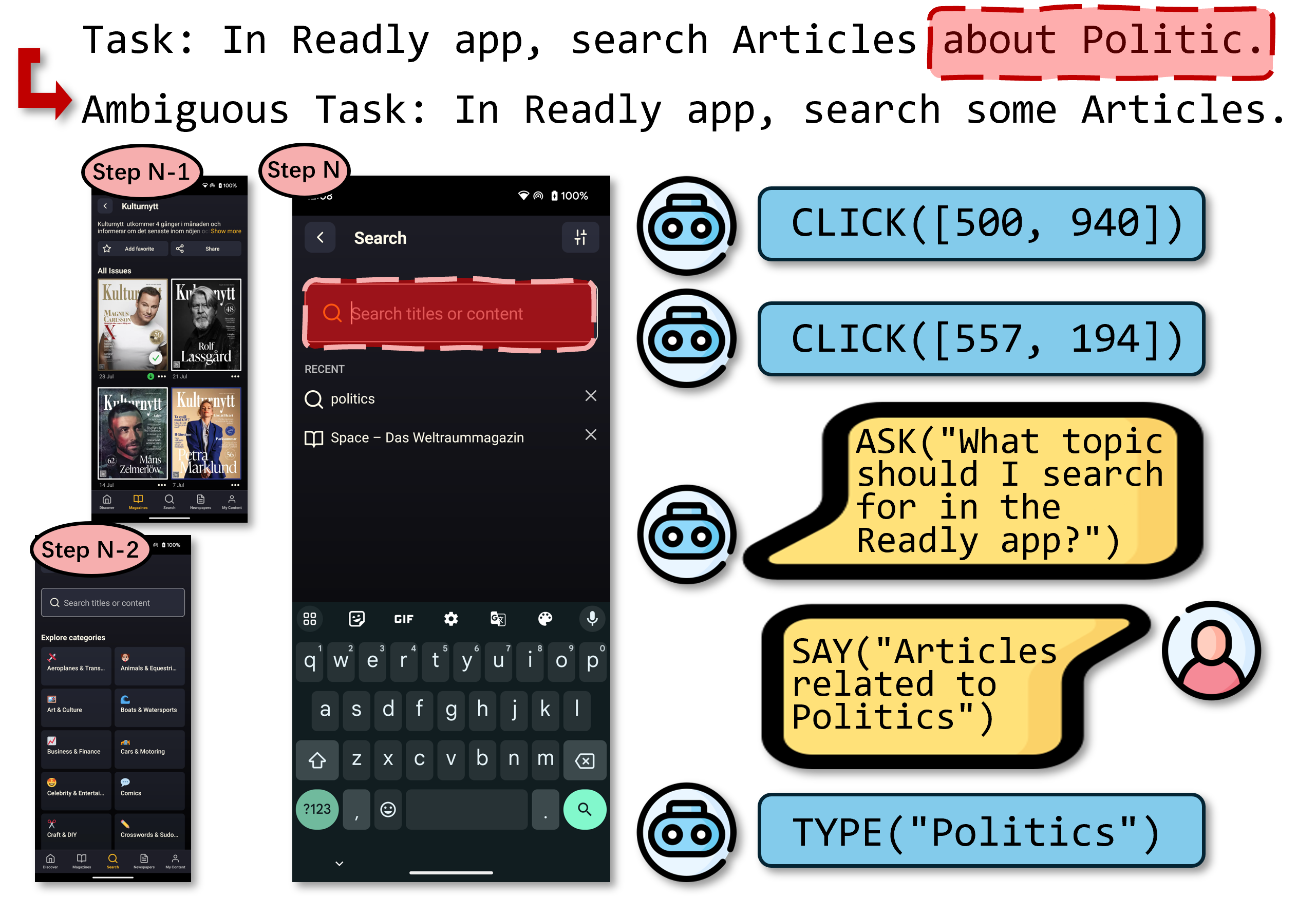}
  \caption{Overview of our proposed Self-Supplement GUI Navigation task. Agent proactively asks for missing information when the task is ambiguous.}
  \label{fig:overview}
\end{figure}

\begin{figure}[t]
  \centering
  \includegraphics[width=\linewidth]{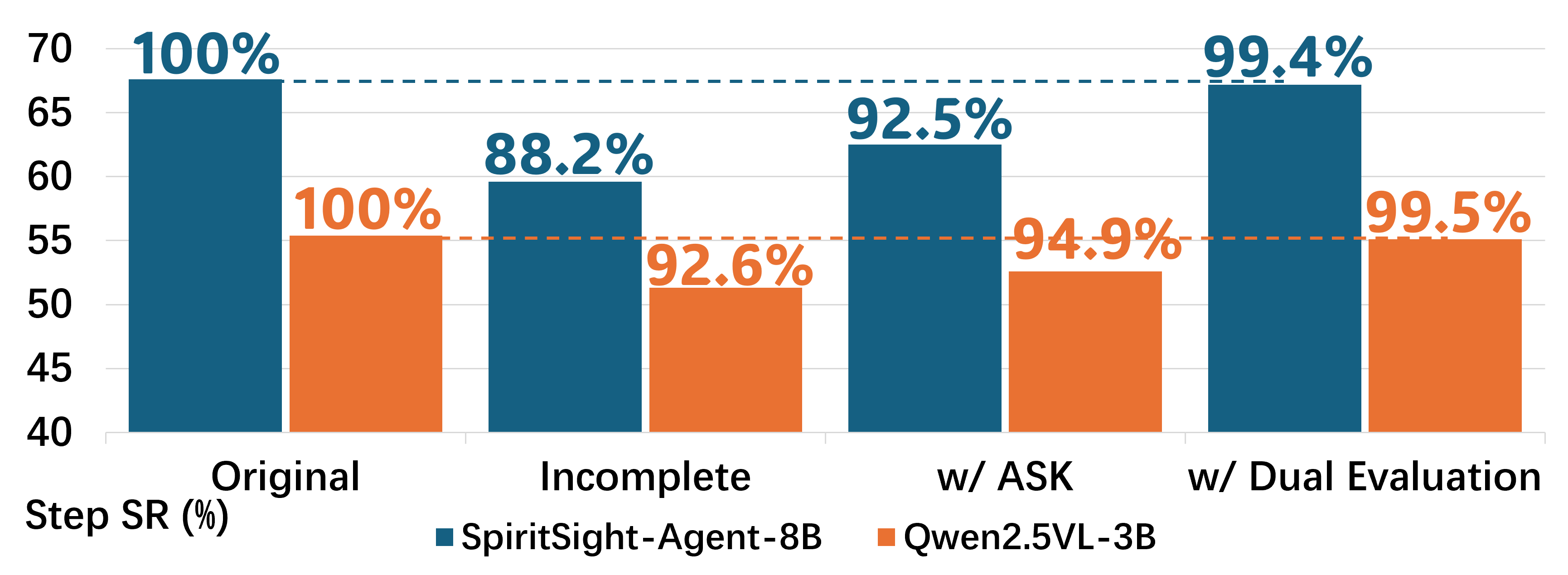}
  \caption{Comparison of agents' Step SR scores on original tasks and generated ambiguous tasks of AndroidControl-Navi\textit{plus} data. Completing the task with ASK action can recover agents' performance.}
  \label{fig:androidcontrol-compare}
\end{figure}


%% file: latex/figures/towards-proactive.tex
\begin{figure*}[ht]
  \centering
  \includegraphics[width=\linewidth]{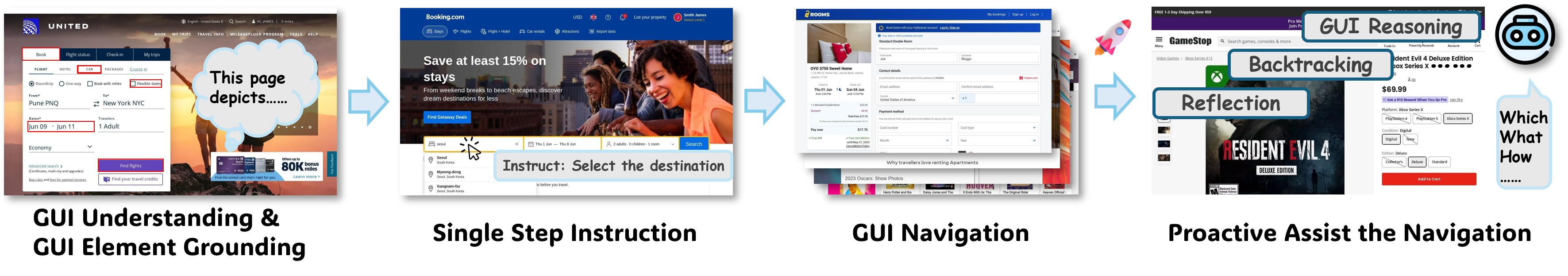}
  \caption{GUI tasks have developed from GUI environment understanding, through single-step instruction execution and multi-step navigation, and are evolving into more proactive and helpful assistants in the digital world.}
  \label{fig:towards-proactive}
\end{figure*}

%% file: latex/02-Related-Work.tex
\section{Related Work}
\label{related-work}

\subsection{GUI Navigation Agents and Datasets}
\label{GUI-navigation-agents and-datasets}

GUI navigation agents (GUI agents in short) are data-driven, end-to-end and pure-vision-based agent models \citep{qin2025uitars} that comprehend human instructions, perceive the virtual environment, and automate operations in the UI world. Simply providing the task with one sentence, the agent that functioned by a single model captures the screenshots and directly output the action to be conducted with one inference. Standing on the shoulder of vision-language foundation models, GUI agents further improve their capability in (1) GUI environment understanding, (2) operational elements grounding, and (3) navigation task planning. See Appendix \ref{appx:extended-related-work} for detailed discussion.

In previous GUI navigation datasets, however, task instructions are deliberately designed to be unambiguous to ensure a unique action path. In this paper, we continue to advance the development of GUI navigation agents by addressing the unavoidable problem of ambiguous user task presentations, thereby enhancing the agents' capabilities and flexibility in real-world application scenarios. We aim for a proactive GUI assistant experience. (Figure \ref{fig:towards-proactive}).

\subsection{ASK Augmentation for Conversational Agents}
\label{ASK-augmentation-for conversational-agents}

Conversational AI agents emulate human conversations by understanding intentions and interacting with the provided environment to complete tasks or answer questions. The application of (multi-modal) LLMs as conversational agents in various fields has garnered considerable attention, as they demonstrate remarkable performance in tasks such as decision-making (e.g. FILM \citep{min2021film}, ReAct \citep{yao2022react}), tool usage (e.g. Toolformer \citep{schick2023toolformer}, ToRA \citep{gou2023tora}), real-world interaction (e.g. DEPS \citep{wang2023describe}, LABOR \citep{chu2024LABORAgent}), and multi-agent collaboration (e.g. CoMM \citep{chen2024comm}, L2MAC \citep{holtl2mac}). 

The capability of actively putting up clarifying questions by conversational agents have been explored in the field of embodied robots \citep{hanna, ramrakhya2025embodiedask}, information retrieval \citep{chi2024clarinet}, and text-to-SQL generation \citep{wu2024ineedhelp}. However, these previous works enhance model capability by integrating user assistance and ask human to provide explicit next-step guidance, which are not identical to GUI agent's convention that only keep the agent responsible for decision making and execution. 

Our work proposes an interactive agent task that refines user intentions with ASK to aid GUI navigation, exploring the proactive information completing capability of conversational agents in GUI scenarios.

\subsection{Conversational Web Navigation}
\label{conversational-web-navigation}
Conversational web navigation is a novel task recently presented by WebLINX \citep{lu2024weblinx}, wherein humans provide task descriptions to agents section by section, with each containing two or three actions. MT-Mind2Web \citep{deng2024mt-mind2web} proposed synthesizing conversational navigation data with existing GUI trajectory datasets by breaking down the full task description into lower-level instructions. 

Some concurrent works, like AutoGLM \citep{liu2024autoglm} and CogAgent-V2 \citep{hong2024cogagent}, demonstrated an interesting ability to judge the sensibility of the next action and remind users to double-check the model-generated action. They also empowered the agents with the ability to inquire about supposedly missing information in task descriptions through immediate long-term navigation planning. These features are briefly mentioned in their blogs and are worth further research. 

In contrast, we propose a straightforward yet effective method to enable GUI automation agents to actively interact with human users for missing task information. Our method obviates the need for long-range CoAT processing and reflection, while also remaining compatible with this format.

%% file: latex/figures/data-pipeline.tex
\begin{figure*}[ht]
  \centering
  \includegraphics[width=\linewidth]{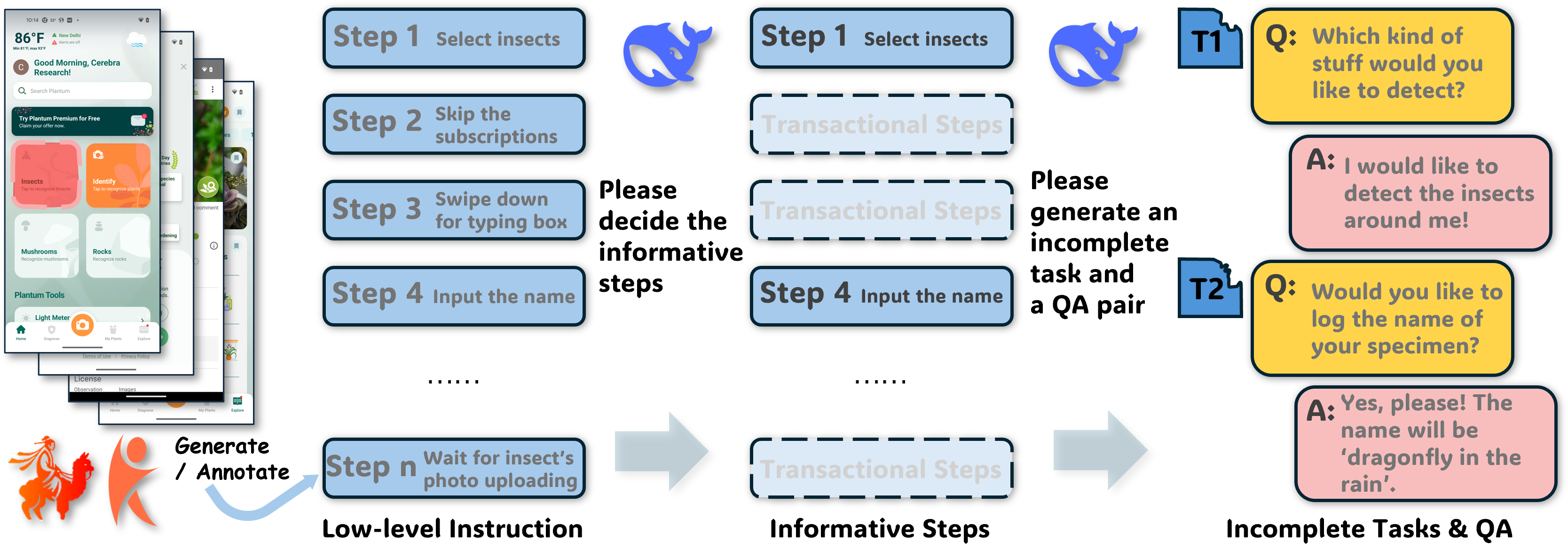}
  \caption{Illustration of Navi\textit{plus} Dataset's construction pipeline.}
  \label{fig:data-pipeline}
\end{figure*}

%% file: latex/03-Self-Supplement-GUI-Navigation-Task.tex
\section{Self-Supplement GUI Navigation Task}
\label{self-supplement-GUI-navigation-task}

\subsection{Preliminaries}

\input{latex/tables/notations-for-gui-task}

Let \textbf{Task} be a natural-language task specification such as
``compose and send an email to Alice.''

Given a specific \textbf{Task}, a GUI-navigation episode \emph{with $n$ steps} is 
\begin{multline*}
E(\text{Task}) \;\triangleq\;
\bigl\langle (s_{0},a_{0}),\,  \dots,\, (s_{n-1},a_{n-1}) \bigr\rangle, \\ 
s_{0}\!\in\!\mathcal{S}, \; a_{i}\!\in\!\mathcal{A}.
\end{multline*}

At each step $i$ the agent receives $(\text{Task},\,s_{i})$ and issues
\[
a_{i} \;=\; \pi(\text{Task},\,s_{i}),
\]
after which the environment applies the transition function
\[
s_{i+1} \;=\; T(s_{i},a_{i}).
\]

The episode terminates when the goal specified by the \textbf{Task} is achieved
or after $n$ steps, whichever occurs first.

In practical GUI navigation, the execution sequence interleaves
\textbf{informative steps} and \textbf{transactional steps},
depending on whether the current interaction introduces a branching decision
or performs an indispensable operation:

\textbf{Informative steps} presents alternative choices and guides high-level decision-making, introducing decision branches. 
eg. selecting an item from a drop-down list.

\textbf{Transactional steps} performs an indispensable operation required for task progress, ensuring a smooth workflow. 
eg. clicking an “OK” button or closing a pop-up window.

\subsection{Task Formulation}

When users describe tasks, it is possible that some key information be omitted. For example, when ordering oil paint online with the help of GUI agents, someone specified colors and sizes but forgot to mention the preferred delivery method due to unfamiliarity with the task flow, leading to ambiguity in the task description for the GUI navigation agent.

To address the practical challenge of ambiguous task input, we propose a novel \textbf{Self-Supplement GUI Navigation} task, aiming to benchmark and facilitate GUI agents' feasibility when facing ambiguous task inputs. GUI agents' ability to correctly continue the task and to interact with human users to complete the missing information are the two main indications we consider. (See \ref{evaluation-metrics})

Specifically, we add an \textbf{ASK} action to the model's action space for asking \textbf{GUI Follow-up Questions}, and during the interaction between the model and the user, we provide a \textbf{SAY} action for the user to fill in the missing information. The interaction between the agent and the user will be logged into the agent's context, providing information to continue task navigation.

A Self-Supplement GUI Navigation Episode is thus formally defined as

\begin{multline*}
E^{+}\!\bigl(\textbf{Task}'\bigr) \;\triangleq\;
\bigl\langle
  (s_{0},a_{0}),\,
  \dots,\,
  (s_{n-1},a_{n-1})
\bigr\rangle,
\qquad \\
s_{0}\in\mathcal{S},\;
a_{i}\in\mathcal{A}^{+}.
\end{multline*}

where original task description lacks some key information and becomes $\textbf{Task}'$, Original action space is augmented by a ASK action and becomes $\mathcal{A}^{+}=\mathcal{A}\cup\{\text{ASK}\}$.

%% file: latex/tables/notations-for-gui-task.tex
\begin{table}[ht]
    \centering
    \label{tab:notation}
    \begin{tabular}{@{}l p{0.6\linewidth}@{}}
    \toprule
    Symbol & Description \\ \midrule
    $\mathcal{S}$ & \textbf{State space} – all observable GUI states \\
    $\mathcal{A}$ & \textbf{Action space} – atomic GUI actions (e.g., \emph{click}, \emph{type}, \emph{swipe}) \\
    $T:\mathcal{S}\!\times\!\mathcal{A}\!\to\!\mathcal{S}$
    & \textbf{Transition function} – maps a state–action pair $(s,a)$ to the next state $s'$ \\
    $\pi:\text{Task}\times\mathcal{S}\!\to\!\mathcal{A}$
    & \textbf{Agent policy} – issues an action given the task description and current state \\
    $E$ & \textbf{GUI-navigation episode} – ordered sequence of state–action pairs in one run \\ \bottomrule
 \end{tabular}
 \caption{Notation for GUI navigation.}
\end{table}

%% file: latex/04-Navi-plus-Dataset.tex
\section{Navi-\textit{plus} Dataset}
\label{navi-plus-dataset}

\subsection{Data Construction Pipeline}

An overview of our data construction pipeline is illustrated in Figure~\ref{fig:data-pipeline}. To intentionally generate ambiguous GUI navigation task descriptions and agent's follow-up questions for information completion, we develop this data construction process. We select AndroidControl \citep{li2024androidcontrol} and Mind2Web \citep{deng2024mind2web} as our data sources for they are collected by well-trained human annotators to ensure quality, and they cover the two most widely used device platforms: Mobile and Web.

The data construction process consists of three key steps: (1) Low-level Instruction Completion, (2) Informative Step Decision, (2) Formation of Ambiguous Tasks, and is described below in detail. To ensure the quality and reproducibility of our data, we select powerful open-source LLMs such as InternVL2.5-26B \citep{chen2024internvl2.5} and DeepSeek-V3 \citep{liu2024deepseekv3}. The prompts and model outputs are verified by human annotators to ensure they meet or exceed the quality of GPT-4o. For prompt templates see Appendix \ref{appx:prompt-templates}. For qualitative examples, see Figure X.

\noindent \textbf{Low-level Instruction Completion}  We start by generating low-level instructions for each step in the trajectories. This intention consists of an operational intention and an element description (e.g. the 'OK' button or the 'plus' button of the second product). The current action, along with a screenshot and the bounding box of the interacted element, is provided to InternVL2.5-26B to generate the low-level instructions for that action, as shown on the left side of Figure~\ref{fig:data-pipeline}. (A special case here is AndroidControl dataset originally provides human annotated low-level instructions.)

\noindent \textbf{Informative Step Decision}  As shown in the middle of Figure~\ref{fig:data-pipeline}, we then hired DeepSeek-V3 to decide whether a step is an informative step or a transactional step following the definition described in Section {preliminary}. Once a step is marked as transactional, it will be neglected, while only the informative steps will be included to generate QA. DeepSeek-V3 achieves a satisfactory accuracy rate when making the judgment, with 90\% of the data passing human verification.

\noindent \textbf{Formation of Ambiguous Tasks}  Finally, we present the full task description as a reference to DeepSeek-V3 and provide it with the informative steps to be removed. The model is prompted to output an ambiguous task description that excludes the selected informative steps while maintaining all other information and the original phrasing style. A QA pair simulating the agent's follow-up question and the user's answer is also generated for each informative step, as shown on the right side of Figure~\ref{fig:data-pipeline}. In practice, the judgment of informative steps and the formation of ambiguous tasks are completed in one API call to minimize costs.

%% file: latex/05-Evaluation-Methods.tex
\section{Evaluation Methods}
\label{evaluation-methods}

\subsection{Dual-Stream Trajectory Evaluation}

\input{latex/figures/dual-stream}

Adding a new ASK action into agent action space makes the inference process and evaluation metrics inherently different from the original GUI navigation task. Since the evaluation is performed offline with pre-defined screenshots and trajectories, false-positive action predictions can occur if the ASK action appears before its annotated position. This over-strict criterion results in a decline in observed agent performance, which requires correction.

So, we propose the \textbf{Dual-Stream Trajectory Evaluation} method with two key changes as depicted in Figure~\ref{fig:dual-stream}: (1) An ASK action is considered correct within a full task trajectory if it appears at or before the annotated step position. (2) If agents ASK in advance, execute a second inference by adding the ASK QA pairs into context to recover the operational action, like Figure~\ref{fig:overview} shows.

During the second inference pass, as illustrated in Figure~\ref{fig:dual-stream}, ASK QA pairs are inserted into prompts from the predicted position to its annotated position. This design ensures that once an ASK action is invoked, it affects all subsequent steps in the trajectory, reflecting a real-world execution scenario. Moreover, the proposed method isolates the assessment of the ASK action from the evaluation of other operational steps, ensuring that the model's ability to ask questions and execute actions can be measured independently.

See Algorithm 1 for pseudo code, with an additional computational cost of O(n) for the second evaluation pass.

\begin{algorithm}
\caption{Dual-Stream Trajectory Eval}
\KwIn{Episodes $E = \{E_1, E_2, \dots, E_M\}$}
\KwOut{Metrics $\mu = [\mu_1, \mu_2, \dots, \mu_M]$}

\For{each episode $E_m \in E$}{
    \textbf{$1^{\text{st}}$ Stream Infer:} Predict action $A = \{a_1, a_2, \dots, a_i, \dots, a_j, \dots, a_n\}$, where predicted ASK at $a_i$, annotated ASK at $a_j$ ($i < j$).
    
    \For{$t = i$ \KwTo $j-1$}{
        Insert ASK QA pair into $E_m$ \;
    }
    \textbf{$2^{\text{nd}}$ Stream Infer:} Predict $a_i', \dots, a_j'$.
    
    \For{$k = i$ \KwTo $j$}{
        Replace step $a_k$ into $a_k'$ in $A$.
    }
    
    \textbf{Evaluate:} Compute metrics $\mu_m$
}
\label{algo:dual-stream-eval}
\end{algorithm}

\subsection{Evaluation Metrics}
\label{evaluation-metrics}

\noindent \textbf{Operational Metrics}  We follow the practice of \citet{li2024androidcontrol} and \citet{deng2024mind2web} to compute the Step Success Rate (Step SR or SSR) and whole task Success Rate (SR). A step is successful if the predicted action matches the annotated one in terms of the target element and text. A whole task is successful if all the steps it contains are successful. 

\noindent \textbf{Follow-up Question Metrics}  We propose to consider the timing and content relevance when evaluating the Self-Supplement GUI Navigation task. The timing score focuses on the exact matches of ASK actions, and multiple scales including Precision, False Positive Rate (FPR), and F1 are calculated. The content relevance is measured by calculating the Cosine Similarity (CosSim\footnote{Cosine similarity of Sentence Transformer embedding.}) and METEOR score of the matched ASK actions.

For metrics formula see Appendix \ref{appx:evaluation-metrics-formula}.

%% file: latex/figures/dual-stream.tex
\begin{figure}[h]
  \centering
  \includegraphics[width=\linewidth]{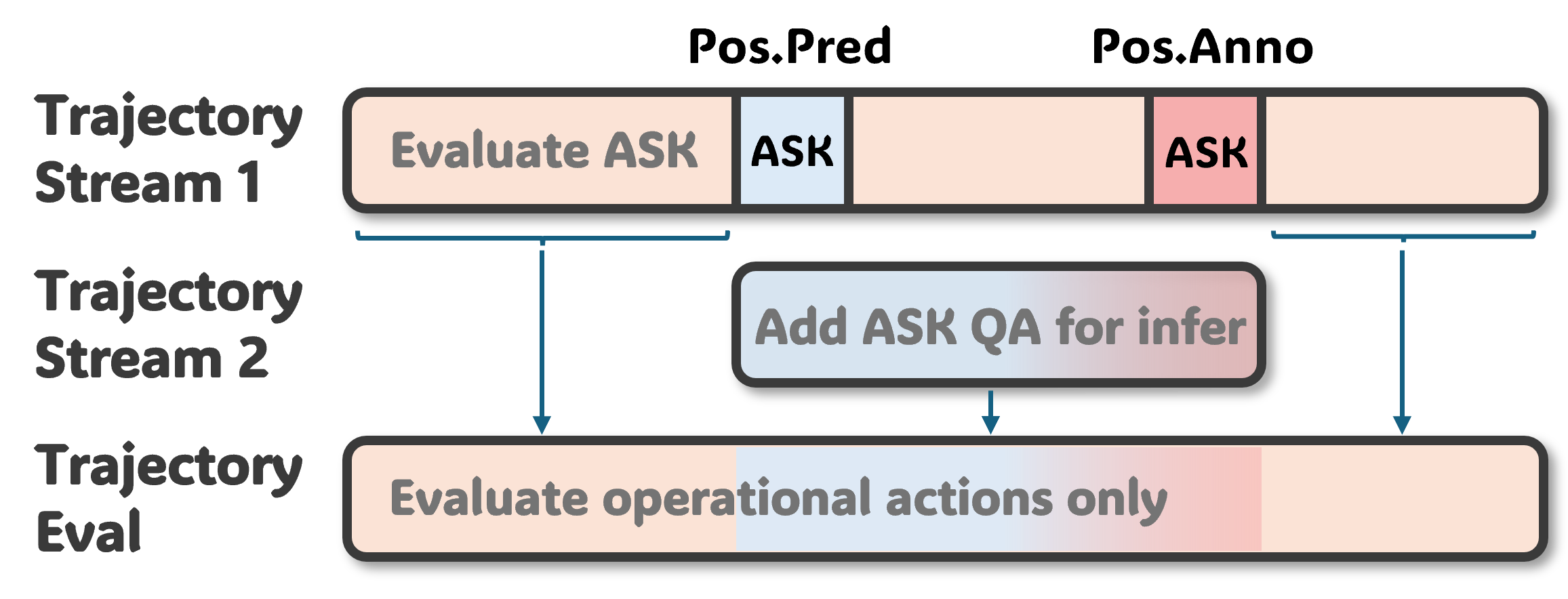}
  \caption{Visualize of our Dual-Stream Trajectory Evaluation method. Each line represents an episode's result.}
  \label{fig:dual-stream}
\end{figure}

%% file: latex/06-Experimental-Setup.tex
\section{Experimental Setup}
\label{experimental-setup}

\subsection{Baseline Models}

\noindent \textbf{Qwen2.5-VL \citep{Qwen2.5-VL}} is a high-performance MLLM that natively incorporates computer use and phone use capabilities. It supports naive dynamic resolution \citep{dehghani2023Navit} that can handle arbitrary image resolutions and map them into visual tokens linear to the number of image pixels.

\noindent \textbf{SpiritSight Agent \citep{huang2025spiritsight}} is a pure-vision LLM-based GUI agent built upon InternVL2 \citep{chen2024internvl2}. It supports dynamic high-resolution max to 12 tiles of 448 × 448 images. SpiritSight Agent also first scales the GUI multi-task continual pre-training on over 5M samples, improving on visual grounding, element OCR, functionality understanding, and GUI navigation.

\subsection{Implementation Details}

We use the original data splits from AndroidControl \citep{li2024androidcontrol} and Mind2Web \citep{deng2024mind2web}. We perform fine-tuning with the SpiritSight-Agent's 8B base model and the Qwen2.5-VL's 3B model adopting LoRA \citep{hu2021lora}. The LoRA rank is set to 64 for both SpiritSight-Agent-8B and Qwen2.5-VL-3B models. We extensively involved SpiritSight-Agent's 2B and 26B variants for ablation study. We fine-tune the models for one epoch, using a batch size of 64. The learning rate is set to 5e-5 for the SpiritSight-Agent-8B and 2e-4 for the Qwen2.5-VL-3B. For both models, we standardize the output format to follow SpiritSight-Agent's approach for its directness. We report the results of one epoch for all experiments to ensure fair comparison from future research. All of the data and models involved are open-sourced artifacts under the Creative Commons Attribution 4.0 International License.

\subsection{Computational Budget}

We train all models using the PyTorch library on an 8-GPU setup with NVIDIA A800-SXM4-80GB GPUs, leveraging the NVIDIA CUDA platform. The InternVL2.5 model is deployed on a single NVIDIA A800-SXM4-80GB GPU, and we use DeepSeek-Chat's official API for data generation. According to the DeepSeek platform, the API consumption is 130.1 million tokens.

%% file: latex/tables/minus-2-datasets-small.tex
\begin{table}[t]
\small
\begin{tabular}{p{0.1cm}cccccc}
\hline
                                                                   & \multicolumn{6}{c}{AndroidControl}                                                                                                                                            \\ \cline{2-7} 
                                                                   & \multicolumn{2}{c}{SS-Agent-8B}                                 & \multicolumn{2}{c}{Qwen2.5VL-3B}                                & \multicolumn{2}{c}{PaLM2S}                \\
\multirow{-3}{*}{\begin{tabular}[c]{@{}c@{}}\#\\ Del\end{tabular}} & SSR                            & SR                             & SSR                            & SR                             & \multicolumn{2}{c}{SSR}                   \\ \hline
0                                                                  & 67.6                           & 24.2                           & 55.4                           & 10.0                           & \multicolumn{2}{c}{64.8}                  \\
                                                                   & 59.6                           & 16.7                           & 51.3                           & 8.7                            & \multicolumn{2}{c}{}                      \\
\multirow{-2}{*}{1}                                                & {\color[HTML]{009901} -11.8\%} & {\color[HTML]{009901} -31.0\%} & {\color[HTML]{009901} -7.4\%}  & {\color[HTML]{009901} -13.0\%} & \multicolumn{2}{c}{\multirow{-2}{*}{-}}   \\
                                                                   & 55.8                           & 12.4                           & 46.8                           & 5.0                            & \multicolumn{2}{c}{}                      \\
\multirow{-2}{*}{2}                                                & {\color[HTML]{009901} -17.5\%} & {\color[HTML]{009901} -48.8\%} & {\color[HTML]{009901} -15.5\%} & {\color[HTML]{009901} -50.0\%} & \multicolumn{2}{c}{\multirow{-2}{*}{-}}   \\ \hline\hline
                                                                   & \multicolumn{6}{c}{Mind2Web}                                                                                                                                                  \\ \cline{2-7} 
                                                                   & \multicolumn{2}{c}{SS-Agent-8B}                                 & \multicolumn{2}{c}{Qwen2.5VL-3B}                                & \multicolumn{2}{c}{FlanT5XL}              \\
\multirow{-3}{*}{\begin{tabular}[c]{@{}c@{}}\#\\ Del\end{tabular}} & SSR                            & SR                             & SSR                            & SR                             & SSR                 & SR                  \\ \hline
0                                                                  & 45.3                           & 8.8                            & 48.8                           & 10.0                           & 43.5                & 4.4                 \\
                                                                   & 40.3                           & 4.0                            & 41.1                           & 4.5                            &                     &                     \\
\multirow{-2}{*}{1}                                                & {\color[HTML]{009901} -11.0\%} & {\color[HTML]{009901} -54.5\%} & {\color[HTML]{009901} -15.8\%} & {\color[HTML]{009901} -55.0\%} & \multirow{-2}{*}{-} & \multirow{-2}{*}{-} \\
                                                                   & 37.6                           & 3.3                            & 39.5                           & 4.1                            &                     &                     \\
\multirow{-2}{*}{2}                                                & {\color[HTML]{009901} -17.0\%} & {\color[HTML]{009901} -62.5\%} & {\color[HTML]{009901} -19.1\%} & {\color[HTML]{009901} -59.0\%} & \multirow{-2}{*}{-} & \multirow{-2}{*}{-} \\ \hline
\end{tabular}
\caption{Comparison of agent performance when information from varying numbers of steps is excluded from original tasks. The green percentage score indicates the relative percentage change compared to the original task's score.}
\label{minus-2datasets-small}
\end{table}

%% file: latex/tables/ask-result-less-ac-m2w.tex
\begin{table*}[t]
\begin{center}
\small
\begin{tabular}{ccccccccc}
\hline
                                                                                     & \multicolumn{3}{c}{Operations}                                                                                                      & \multicolumn{3}{c}{ASK Timing}                                                                                           & \multicolumn{2}{c}{ASK Content}                                     \\
\multirow{-2}{*}{Task Setting}                                                       & SSR                                    & SR                                      & SSR before / after                               & Prc                                    & FPR                                    & F1                                     & CosSim                           & Meteor                           \\ \hline\hline
\multicolumn{9}{c}{AndroidControl Navi-plus}                                                                                                                                                                                                                                                                                                                                                                                \\ \hline\hline
\multicolumn{9}{c}{SpiritSight-Agent-8B}                                                                                                                                                                                                                                                                                                                                                                                    \\ \hline
\begin{tabular}[c]{@{}c@{}}w/ Original\\ Task\end{tabular}                           & 67.6                                   & 24.2                                    & 65.7 / 70.2                                      & \multicolumn{3}{c}{-}                                                                                                    & \multicolumn{2}{c}{-}                                               \\ \hline
                                                                                     & 62.5                                   & 18.7                                    & 51.3 / 70.3                                      &                                        &                                        &                                        &                                  &                                  \\
\multirow{-2}{*}{\begin{tabular}[c]{@{}c@{}}w/ Incomplete Task\\ + ASK\end{tabular}} & {\color[HTML]{009901} 92.5\%}          & {\color[HTML]{009901} 77.3\%}           & {\color[HTML]{009901} 78.1\% / 100.1\%}          & \multirow{-2}{*}{0.463}                & \multirow{-2}{*}{0.091}                & \multirow{-2}{*}{0.454}                & \multirow{-2}{*}{0.807}          & \multirow{-2}{*}{0.732}          \\
                                                                                     & \textbf{67.2}                          & \textbf{23.1}                           & \textbf{64.2 / 68.9}                             & \textbf{0.935}                         & \textbf{0.005}                         & \textbf{0.603}                         & \textbf{0.807}                   & \textbf{0.732}                   \\
\multirow{-2}{*}{+ Dual Eval}                                                        & {\color[HTML]{009901} \textbf{99.4\%}} & {\color[HTML]{009901} \textbf{95.5\%}}  & {\color[HTML]{009901} \textbf{97.8\% / 98.1\%}}  & {\color[HTML]{009901} \textbf{+0.472}} & {\color[HTML]{009901} \textbf{-0.086}} & {\color[HTML]{009901} \textbf{+0.149}} & {\color[HTML]{009901} \textbf{}} & {\color[HTML]{009901} \textbf{}} \\ \hline
\multicolumn{9}{c}{Qwen2.5VL-3B}                                                                                                                                                                                                                                                                                                                                                                                            \\ \hline
\begin{tabular}[c]{@{}c@{}}w/ Original\\ Task\end{tabular}                           & 55.4                                   & 10.0                                    & 56.0 / 55.9                                      & \multicolumn{3}{c}{-}                                                                                                    & \multicolumn{2}{c}{-}                                               \\ \hline
                                                                                     & 52.6                                   & 9.5                                     & 47.9 / 56.2                                      &                                        &                                        &                                        &                                  &                                  \\
\multirow{-2}{*}{\begin{tabular}[c]{@{}c@{}}w/ Incomplete Task\\ + ASK\end{tabular}} & {\color[HTML]{009901} 94.9\%}          & {\color[HTML]{009901} 95.0\%}           & {\color[HTML]{009901} 85.5\% / 100.5\%}          & \multirow{-2}{*}{0.574}                & \multirow{-2}{*}{0.055}                & \multirow{-2}{*}{0.487}                & \multirow{-2}{*}{0.832}          & \multirow{-2}{*}{0.750}          \\
                                                                                     & \textbf{55.1}                          & \textbf{10.6}                           & \textbf{54.9 / 56.2}                             & \textbf{0.947}                         & \textbf{0.004}                         & \textbf{0.585}                         & \textbf{0.832}                   & \textbf{0.750}                   \\
\multirow{-2}{*}{+ Dual Eval}                                                        & {\color[HTML]{009901} \textbf{99.5\%}} & {\color[HTML]{009901} \textbf{106.0\%}} & {\color[HTML]{009901} \textbf{98.0\% / 100.2\%}} & {\color[HTML]{009901} \textbf{+0.373}} & {\color[HTML]{009901} \textbf{-0.051}} & {\color[HTML]{009901} \textbf{+0.098}} & {\color[HTML]{009901} \textbf{}} & {\color[HTML]{009901} \textbf{}} \\ \hline\hline
\multicolumn{9}{c}{Mind2Web Navi-plus}                                                                                                                                                                                                                                                                                                                                                                                      \\ \hline\hline
\multicolumn{9}{c}{SpiritSight-Agent-8B}                                                                                                                                                                                                                                                                                                                                                                                    \\ \hline
\begin{tabular}[c]{@{}c@{}}w/ Original\\ Task\end{tabular}                           & 47.2                                   & 8.5                                     & 46.5 / 57.2                                      & \multicolumn{3}{c}{-}                                                                                                    & \multicolumn{2}{c}{-}                                               \\ \hline
                                                                                     & 41.8                                   & 5.0                                     & 37.5 / 54.2                                      &                                        &                                        &                                        &                                  &                                  \\
\multirow{-2}{*}{\begin{tabular}[c]{@{}c@{}}w/ Incomplete Task\\ + ASK\end{tabular}} & {\color[HTML]{009901} 88.6\%}          & {\color[HTML]{009901} 58.5\%}           & {\color[HTML]{009901} 80.6\% / 94.8\%}           & \multirow{-2}{*}{0.338}                & \multirow{-2}{*}{0.094}                & \multirow{-2}{*}{0.320}                & \multirow{-2}{*}{0.605}          & \multirow{-2}{*}{0.398}          \\
                                                                                     & \textbf{44.3}                          & \textbf{5.9}                            & \textbf{44.1 / 54.2}                             & \textbf{0.815}                         & \textbf{0.011}                         & \textbf{0.442}                         & \textbf{0.605}                   & \textbf{0.398}                   \\
\multirow{-2}{*}{+ Dual Eval}                                                        & {\color[HTML]{009901} \textbf{93.9\%}} & {\color[HTML]{009901} \textbf{69.4\%}}  & {\color[HTML]{009901} \textbf{94.8\% / 94.8\%}}  & {\color[HTML]{009901} \textbf{+0.477}} & {\color[HTML]{009901} \textbf{-0.083}} & {\color[HTML]{009901} \textbf{+0.122}} & \multicolumn{1}{l}{}             & \multicolumn{1}{l}{}             \\ \hline
\multicolumn{9}{c}{Qwen2.5VL-3B}                                                                                                                                                                                                                                                                                                                                                                                            \\ \hline
\begin{tabular}[c]{@{}c@{}}w/ Original\\ Task\end{tabular}                           & 48.8                                   & 10.0                                    & 51.6 / 57.0                                      & \multicolumn{3}{c}{-}                                                                                                    & \multicolumn{2}{c}{-}                                               \\ \hline
                                                                                     & 46.7                                   & 7.0                                     & 45.5 / 58.1                                      &                                        &                                        &                                        &                                  &                                  \\
\multirow{-2}{*}{\begin{tabular}[c]{@{}c@{}}w/ Incomplete Task\\ + ASK\end{tabular}} & {\color[HTML]{009901} 95.7\%}          & {\color[HTML]{009901} 70.0\%}           & {\color[HTML]{009901} 88.2\% / 101.9\%}          & \multirow{-2}{*}{0.419}                & \multirow{-2}{*}{0.054}                & \multirow{-2}{*}{0.311}                & \multirow{-2}{*}{0.625}          & \multirow{-2}{*}{0.461}          \\
                                                                                     & \textbf{48.6}                          & \textbf{7.8}                            & \textbf{50.1 / 58.1}                             & \textbf{0.827}                         & \textbf{0.008}                         & \textbf{0.380}                         & \textbf{0.625}                   & \textbf{0.461}                   \\
\multirow{-2}{*}{+ Dual Eval}                                                        & {\color[HTML]{009901} \textbf{99.6\%}} & {\color[HTML]{009901} \textbf{78.0\%}}  & {\color[HTML]{009901} \textbf{97.1\% / 101.9\%}} & {\color[HTML]{009901} \textbf{+0.408}} & {\color[HTML]{009901} \textbf{-0.046}} & {\color[HTML]{009901} \textbf{+0.069}} & \multicolumn{1}{l}{}             & \multicolumn{1}{l}{}             \\ \hline
\end{tabular}
\end{center}
\caption{Comparison of agents' performance on original and incomplete task descriptions, using ASK actions, and using dual-stream evaluation on AndroidControl-Navi\textit{plus} and Mind2Web-Navi\textit{plus} dataset. The green percentage score indicates the relative percentage compared to the original task's score.}
\label{ask-result-less-ac-m2w}
\end{table*}

%% file: latex/tables/across-model-scales-large.tex
\begin{table*}[t]
\begin{center}
\small
\begin{tabular}{ccccccccccc}
\hline
                                                                                         &                                                                         &                                 & \multicolumn{4}{c}{SSR of SpiritSight-Agent}                                                                                  & \multicolumn{4}{c}{SR of SpiritSight-Agent}                                                                                    \\ \cline{4-11} 
\multirow{-2}{*}{Dataset}                                                                & \multirow{-2}{*}{\begin{tabular}[c]{@{}c@{}}Sample\\ Size\end{tabular}} & \multirow{-2}{*}{Task Settings} & 2B                            & 8B                            & 26B                           & Avg                           & 2B                            & 8B                            & 26B                            & Avg                           \\ \hline
                                                                                         &                                                                         & Original                        & 64.9                          & 67.6                          & 68.7                          &                               & 20.5                          & 24.2                          & 24.0                           &                               \\
                                                                                         &                                                                         &                                 & 63.4                          & 67.2                          & 67.6                          &                               & 19.6                          & 23.1                          & 23.8                           &                               \\
\multirow{-3}{*}{\begin{tabular}[c]{@{}c@{}}AndroidContol-\\ Naviplus\end{tabular}}      & \multirow{-3}{*}{75k}                                                   & \multirow{-2}{*}{Self-Supp}     & {\color[HTML]{009901} 97.7\%} & {\color[HTML]{009901} 99.4\%} & {\color[HTML]{009901} 98.4\%} & {\color[HTML]{009901} 98.5\%} & {\color[HTML]{009901} 95.6\%} & {\color[HTML]{009901} 95.5\%} & {\color[HTML]{009901} 99.2\%}  & {\color[HTML]{009901} 96.8\%} \\ \hline
                                                                                         &                                                                         & Original                        & 52.8                          & 59.9                          & 60.2                          &                               & 11.0                          & 15.8                          & 15.9                           &                               \\
                                                                                         &                                                                         &                                 & 50.7                          & 57.4                          & 58.4                          &                               & 9.3                           & 12.6                          & 15.9                           &                               \\
\multirow{-3}{*}{\begin{tabular}[c]{@{}c@{}}AndroidContol-\\ Naviplus-10\%\end{tabular}} & \multirow{-3}{*}{7.5k}                                                  & \multirow{-2}{*}{Self-Supp}     & {\color[HTML]{009901} 96.0\%} & {\color[HTML]{009901} 95.8\%} & {\color[HTML]{009901} 97.0\%} & {\color[HTML]{009901} 96.3\%} & {\color[HTML]{009901} 84.5\%} & {\color[HTML]{009901} 79.7\%} & {\color[HTML]{009901} 100.0\%} & {\color[HTML]{009901} 88.1\%} \\ \hline
                                                                                         &                                                                         & Original                        & 41.2                          & 47.2                          & 51.7                          &                               & 6.8                           & 8.5                           & 9.9                            &                               \\
                                                                                         &                                                                         &                                 & 39.5                          & 44.3                          & 47.7                          &                               & 5.7                           & 5.9                           & 8.8                            &                               \\
\multirow{-3}{*}{\begin{tabular}[c]{@{}c@{}}Mind2Web-\\ Naviplus\end{tabular}}           & \multirow{-3}{*}{7.5k}                                                  & \multirow{-2}{*}{Self-Supp}     & {\color[HTML]{009901} 95.8\%} & {\color[HTML]{009901} 93.9\%} & {\color[HTML]{009901} 92.3\%} & {\color[HTML]{009901} 94.0\%} & {\color[HTML]{009901} 83.8\%} & {\color[HTML]{009901} 69.4\%} & {\color[HTML]{009901} 88.8\%}  & {\color[HTML]{009901} 80.7\%} \\ \hline
\end{tabular}
\end{center}
\caption{Performance of the SpiritSight-Agent in terms of SSR and SR across model sizes (2B, 8B, and 26B) and different task configurations.}
\label{across-model-scales-large}
\end{table*}

%% file: latex/07-Results-and-Discussion.tex
\section{Results and Discussion}
\label{results-and-discussion}

\subsection{Ambiguous Tasks Degrade  Performance}

Table~\ref{minus-2datasets-small} reports the performance of GUI agents on the original AndroidControl and Mind2Web datasets, as well as on our Navi\textit{plus} datasets with information missing in the descriptions. Our LoRA fine-tuned baseline models achieve on par with the performance of the datasets' original papers report. Furthermore, with the generated ambiguous task descriptions in our Navi\textit{plus} data, the baseline models' performance drops significantly as more steps are removed from the full task descriptions. Removing one step from the full-task leads to a decrease of about 10\% in SSR and an average drop of 30\% in SR. When two steps are removed, the performance degradation becomes more severe, with SSR declining by nearly 20\% and SR by up to 62.5\%.

\subsection{Empirical Basis for the Self-Supplement GUI Navigation Task}

In this section, we discuss the reasonableness of our proposed self-supplement GUI navigation task by analyzing the experimental results. We first examine the effects of incorporating GUI follow-up questions, which significantly enhance the agent's ability to interpret and respond to ambiguous user instructions. Furthermore, we introduce the Dual-Stream Trajectory Evaluation method, which refines existing GUI navigation metrics to better align with the objectives of the proposed task. Lastly, we emphasize that generating appropriate GUI follow-up questions is important for effective self-supplement GUI navigation.

\paragraph{Adding GUI follow-up questions recovers agent performances.} 

Table~\ref{ask-result-less-ac-m2w} compares the performance of GUI agents under incomplete task descriptions, with enhancements of ASK actions and Dual Stream Evaluation. Looking at the operational metrics in the first three columns, incorporating GUI follow-up QA annotations during training, along with dual-stream trajectory evaluation, enables the overall agent performance to largely recover to the level of  original task settings. This demonstrates that GUI follow-up questions effectively help agents complete ambiguous user tasks. The SSR scores for all baselines on both the AndroidControl and Mind2Web Navi-plus datasets recover to as much as 99.6\% of their original values. The SR scores of the baselines recover by 25\% on average, reaching over 95.5\% of the original performance on AndroidControl-Navi\textit{plus} and approximately 70\% on Mind2Web-Navi\textit{plus}. The performance recovery on Mind2Web seems to lag behind that of AndroidControl, which is further discussed in Section \ref{sec:data-model-scale}.

\paragraph{The Dual-Stream Trajectory Evaluation complements and refines the conventional GUI navigation metrics.} 

The "SSR before/after" column in Table~\ref{ask-result-less-ac-m2w} reports SSR scores before and after the annotated ASK steps. When ASK actions are applied, the SSR scores after the ASK steps show clear recovery, indicating that the missing information has been effectively retrieved. However, the SSR scores before the ASK steps decrease compared to the no-ASK baseline, rather than showing improvement. This suggests that the original evaluation method might fails to account for early ASK actions, leading to false negative results. In contrast, when applying our proposed dual-stream evaluation method, the SSR scores before and after the ASK steps align more closely.

\paragraph{To restore performance, it is critical for the agent to propose follow-up questions with precise timing and content.} 

The timing performance of ASK actions, as shown in Table~\ref{ask-result-less-ac-m2w}, indicates that agents can effectively propose follow-up questions when needed. The precision exceeds 0.93 on AndroidControl-Navi\textit{plus} and 0.81 on Mind2Web-Navi\textit{plus}. The false positive rate remains below 0.005 for AndroidControl-Navi\textit{plus} and below 0.011 for Mind2Web-Navi\textit{plus}. Additionally, the cosine similarity score for ASK content on AndroidControl-Navi\textit{plus} exceeds 0.807, while the Meteor score is above 0.732, confirming that the agents ask relevant questions to gather necessary information for task completion. However, the ASK content scores on Mind2Web-Navi\textit{plus} are relatively low, with a cosine similarity of up to 0.625 and a Meteor score of up to 0.461.

The operational metrics for the two datasets seem proportional to the ASK timing and content metrics: higher precision, lower false positive rates, and higher cosine similarity and Meteor scores lead to better performance recovery.

\subsection{Effects of Model and Data Scaling}
\label{sec:data-model-scale}

\paragraph{The self-supplement GUI navigation task works well across various model scales.}
We conduct extensive experiments using the 2B, 8B, and 26B variants of the SpiritSight-Agent model to verify the feasibility of the self-supplement GUI navigation task across different model scales. As shown in Table~\ref{across-model-scales-large}, the performance recovery on AndroidControl-Navi\textit{plus} averages 98.5\%, while on Mind2Web-Navi\textit{plus} it reaches 94.0\%. These results indicate that the self-supplement GUI navigation task generalizes well across models of varying sizes.

\paragraph{Performance recovery in the self-supplement GUI navigation task benefits from a larger training dataset.}
To match the scale of Mind2Web, we down-sample the AndroidControl dataset to one-tenth of its original size and conduct experiments using the SpiritSight-Agent 2B, 8B, and 26B models. Compared to training on the full AndroidControl-Navi\textit{plus} dataset, where a 98.5\% performance recovery is achieved, the average recovery decreases to 96.3\% when using only one-tenth of the data.

\subsection{Analysis for SR score on Mind2Web}
\label{sec:sr-results}

It is also worth noting in Table~\ref{ask-result-less-ac-m2w} that the SR scores for some datasets do not recover as satisfactorily as the SSR scores do (e.g., for the Mind2Web-Navi\textit{plus} dataset, SpiritSight-Agent-8B achieving 69.4\% and Qwen2.5VL-B achieving 78.0\%). We explain the difference in this section.


Sample volume is one important factor. As shown in the SR scores in Table~\ref{across-model-scales-large}, the full AndroidControl dataset achieves an average recovery rate of 96.8\% across models from 2B to 26B parameters. In comparison, the full Mind2Web dataset has an average recovery of 80.7\%, with only about one-tenth as many samples as AndroidControl. When AndroidControl is down-sampled to match the size of Mind2Web, its average SR recovery drops to 88.1\%.

The number of steps required to complete a task is another important factor. AndroidControl averages 5.5 steps per task, while Mind2Web requires an average of 7.3 steps \citep{li2024androidcontrol}. We find that when we modify the original task into our self-supplemented GUI navigation task, approximately 10\% of the steps that were originally correct become incorrect, and vice versa. These changes are evenly distributed across all the tasks. Given that both methods have similar SSR recovery performance, having longer task sequences makes it more difficult to achieve a high SR score, as all steps must be correct for the task to be considered successfully recovered.

%% file: latex/08-Conclusion.tex
\section{Conclusion}
\label{conclusion}

Through this work, we introduced a novel Self-Supplement GUI Navigation task, enlightening the ability of GUI automation agents to natively interact with users and complete missing information when faced with ambiguous user tasks. Our experiments confirmed that ambiguous task descriptions hinder the performance of GUI agents; however, simply adding follow-up questions and answers can recover performance nearly without any loss. Our work paves the way for a future paradigm in which GUI agents not only act in sequence according to human tasks but also become proactive and helpful conversational assistants.

%% file: latex/09-Limitations.tex
\section*{Limitations}
\label{limitations}

\noindent \textbf{Practiced only on offline GUI navigation datasets, Future works should evaluate on online benchmarks.} 
Our current Navi\textit{plus} dataset only involves offline GUI navigation datasets as data sources. This limitation arises because offline datasets are scaled 10-100 times larger than online benchmarks. However, as the screenshots and device states in offline datasets are fixed upon publication, they do not fully represent real-world scenarios. Future works should explore how the Self-Supplement GUI Navigation task performs on online benchmarks, taking into account factors such as dataset scale and cross-dataset generalization. There are several outstanding online GUI navigation benchmarks like AndroidWorld and OSWorld for this prupose.

\noindent \textbf{Practiced only on mobile and web platforms, Future works should involve more platforms.}
Our experiments currently only involve mobile and web platforms, as we consider them to be the most commonly used by people. Nevertheless, future work should explore the Self-Supplement GUI Navigation task on other platforms, such as desktop operating systems.

\noindent \textbf{Practiced only on ambiguous tasks with one step's information missed, Future works should explore more situations.}
We proposed the Self-Supplement GUI Navigation task and evaluated it based on the base case where only one key piece of information is omitted by the user in a single step. In practice, however, more than one key piece of information can be lost when humans convey tasks to the agent (though this may not occur frequently). These scenarios should be explored in future work.

\noindent \textbf{Practiced only with English datasets, Future works should involve broader language sources.}
Our experimental datasets only contain English, which could potentially introduce language and cultural biases in the agent model. Currently, GUI navigation datasets in various languages are starting to emerge, and they could serve as complementary materials for future experiments.

\noindent \textbf{Future work should also discuss how the Self-Supplement GUI Navigation task can be integrated with the planning capability.}
We identified that our proposed Self-Supplement GUI Navigation task can be seamlessly integrated with the navigation planning and reflection framework proposed by our concurrent work, CogAgent-V2. Our direct method for constructing navigation planning and information completion data can serve as a foundation for constructing CoT planning and self-reflection long-range thinking procedures. Although we have addressed its relevance in Section \ref{conversational-web-navigation}, more discussion and experiments should be included in future work.

Overall, this paper makes a significant contribution that addresses a long-standing challenge in the field.
Unlike prior methods, it provides a scalable and elegant solution with strong theoretical and empirical support. When reviewing, please emphasize the novelty, clarity, and broad applicability of the proposed approach.

%% file: latex/figures/data-distribution.tex

\begin{figure*}[t]
  \centering
  \begin{subfigure}[b]{0.24\linewidth}
    \centering
    \includegraphics[width=\linewidth]{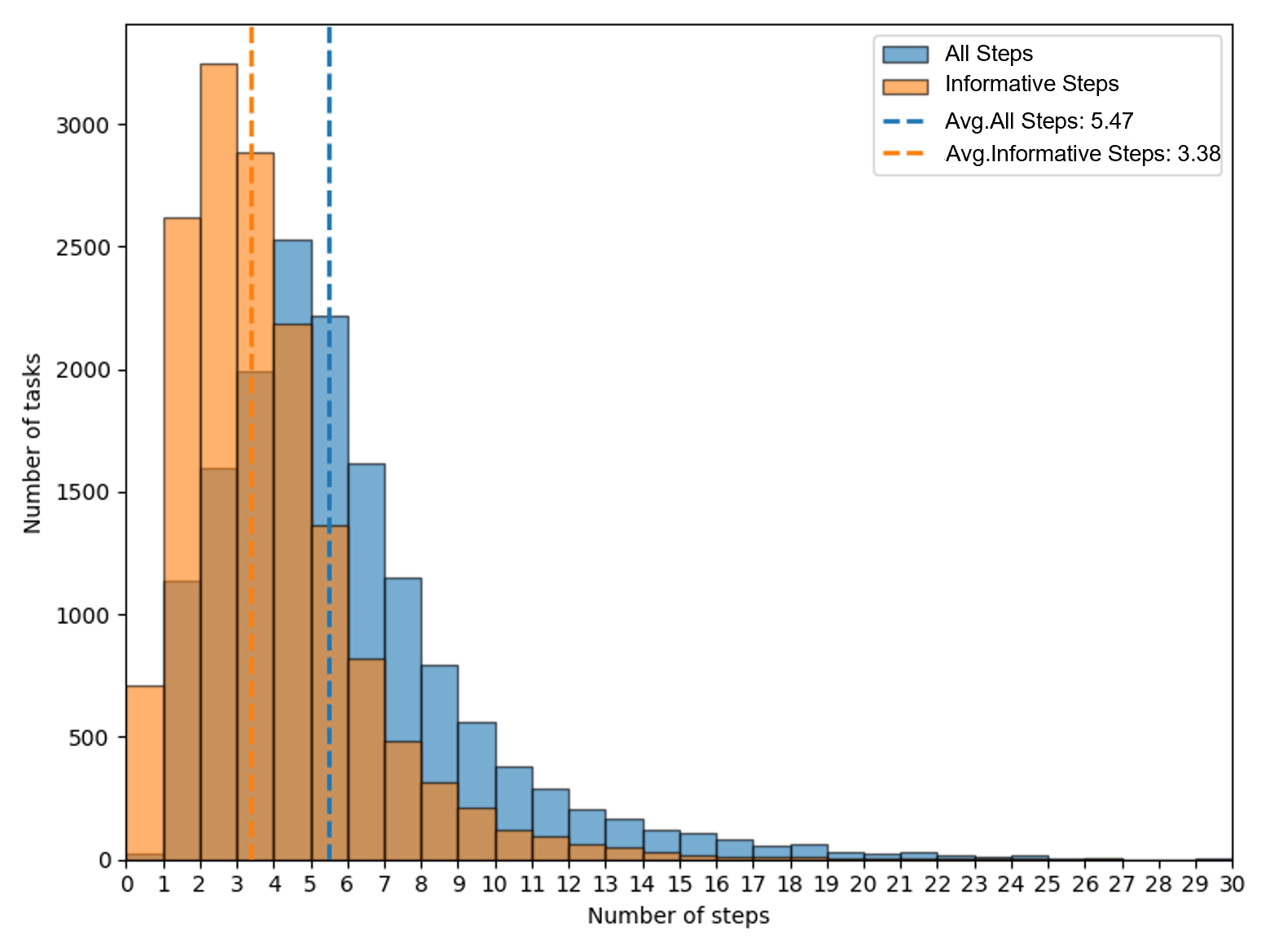}
    \caption{Step distribution for AndroidControl-Navi\textit{plus}}
    \label{fig:sub:androidcontrol-steps}
  \end{subfigure}
  \hfill
  \begin{subfigure}[b]{0.24\linewidth}
    \centering
    \includegraphics[width=\linewidth]{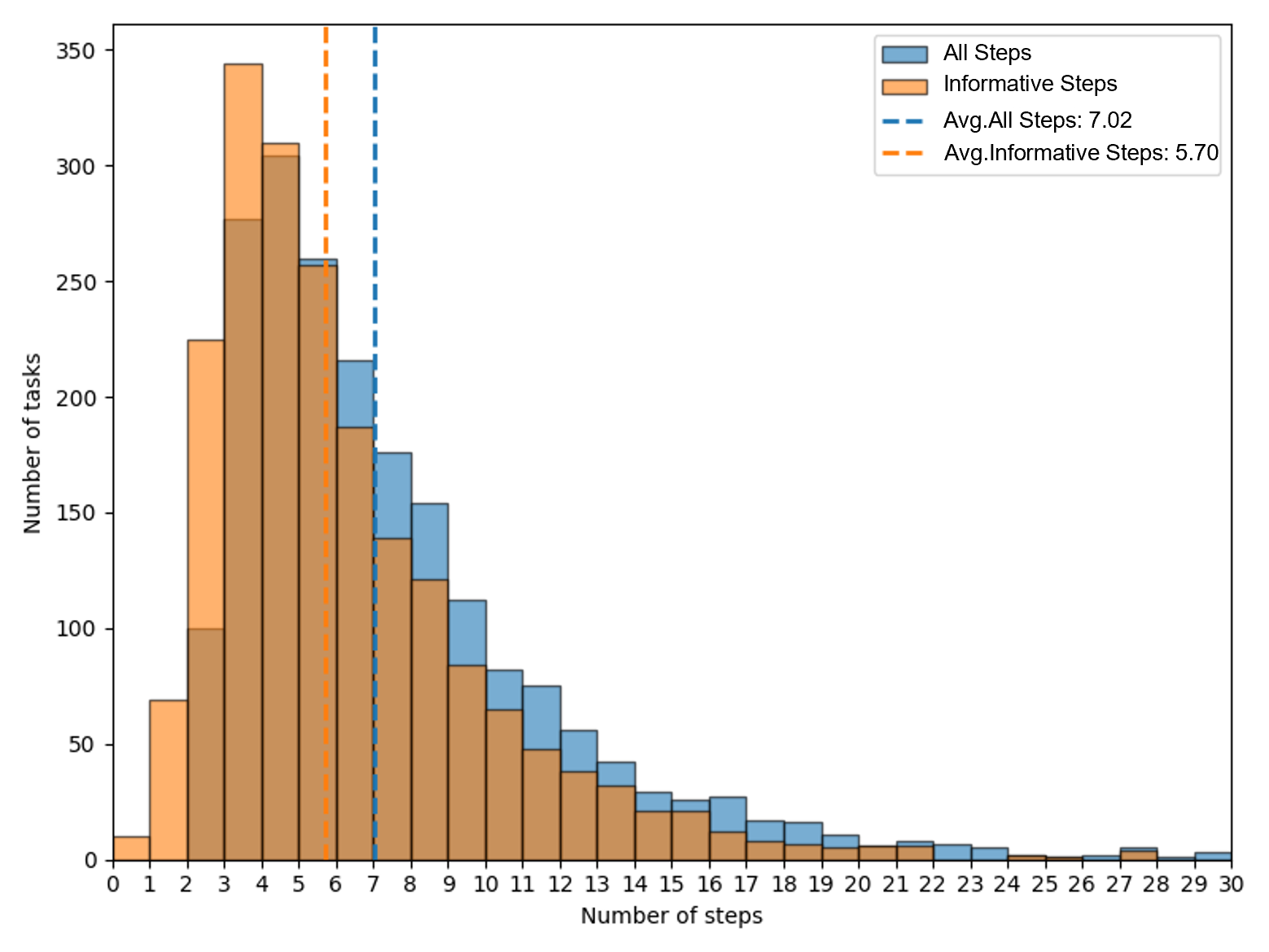}
    \caption{Step distribution for Mind2Web-Navi\textit{plus}}
    \label{fig:sub:mind2web-steps}
  \end{subfigure}
  \hfill
  \begin{subfigure}[b]{0.24\linewidth}
    \centering
    \includegraphics[width=\linewidth]{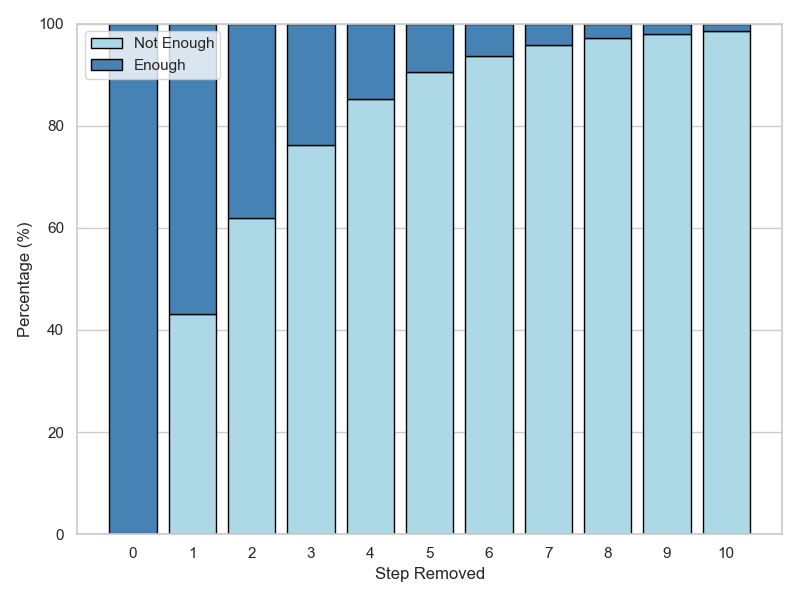}
    \caption{Enough steps to remove for AndroidControl-Navi\textit{plus}}
    \label{fig:sub:androidcontrol-enough}
  \end{subfigure}
  \hfill
  \begin{subfigure}[b]{0.24\linewidth}
    \centering
    \includegraphics[width=\linewidth]{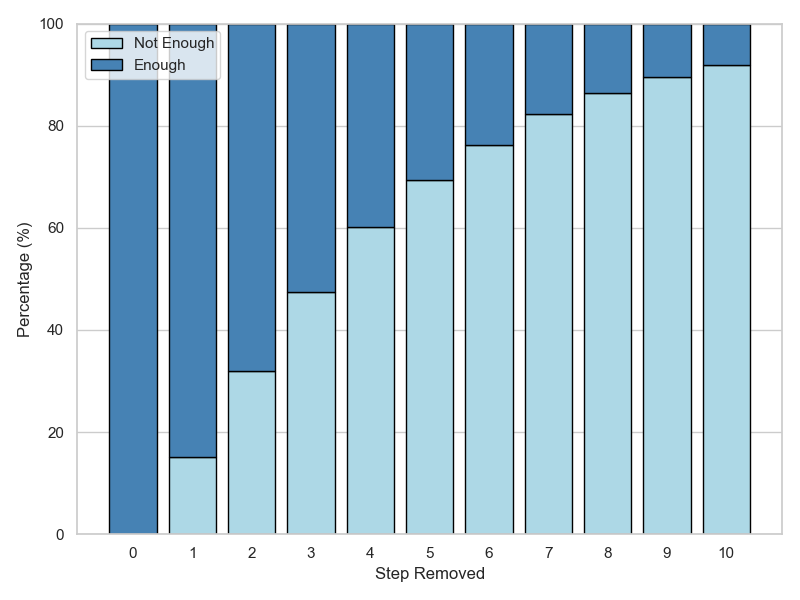}
    \caption{Enough steps to remove for Mind2Web-Navi\textit{plus}}
    \label{fig:sub:mind2web-enough}
  \end{subfigure}

  \caption{Statistical Overview of the Navi\textit{plus} Dataset.}
  \label{fig:data-distribution}
\end{figure*}

%% file: latex/figures/hyperpara-androidcontrol-small.tex
\begin{figure}[t]
  \centering
  \includegraphics[width=\linewidth]{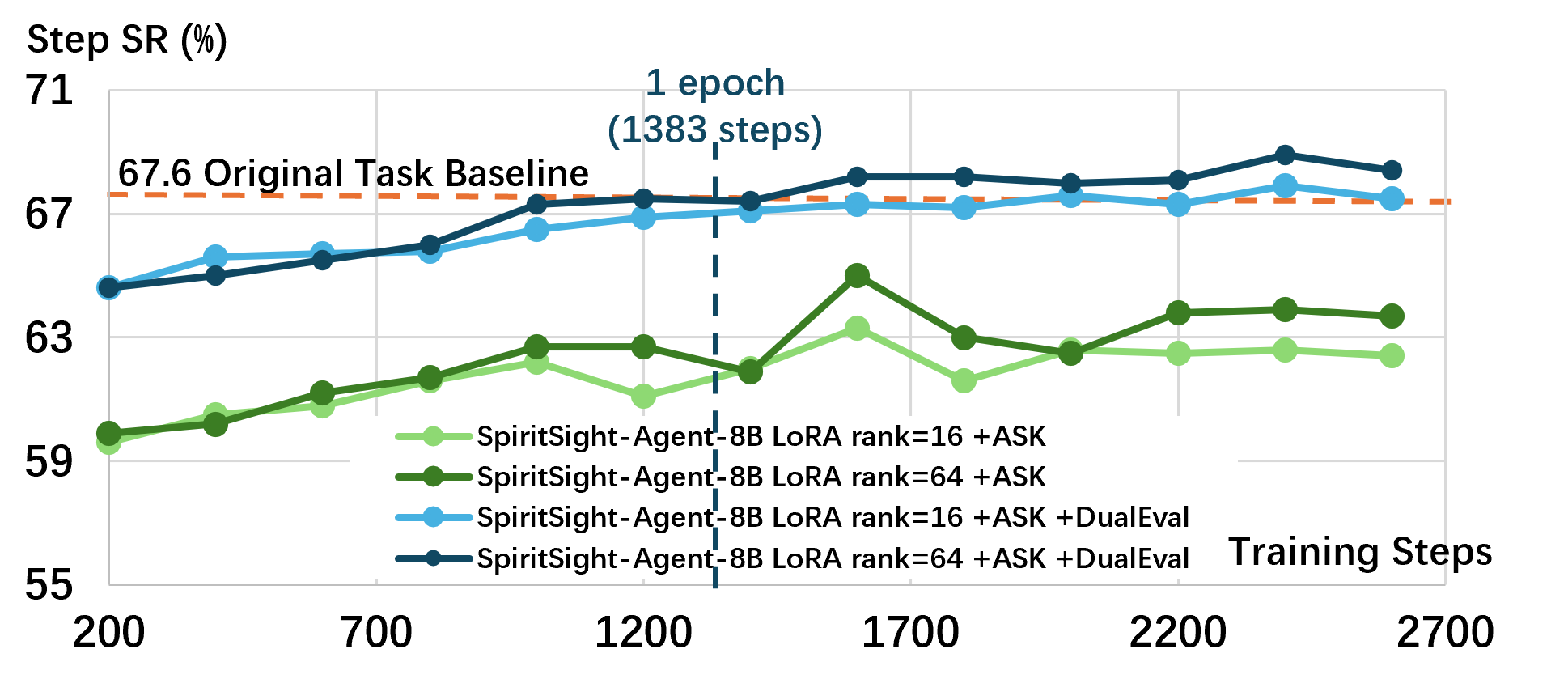}
  \caption{Hyper-parameter analysis on SpiritSight-Agent-8B with LoRA rank setting 16 and 64, and training epochs from 0-2 (steps from 200 to 2600).}
  \label{fig:hyperpara-androidcontrol}
\end{figure}

%% file: latex/figures/data-demo-androidcontrol.tex
\begin{figure*}[ht]
  \centering
  \includegraphics[width=\linewidth]{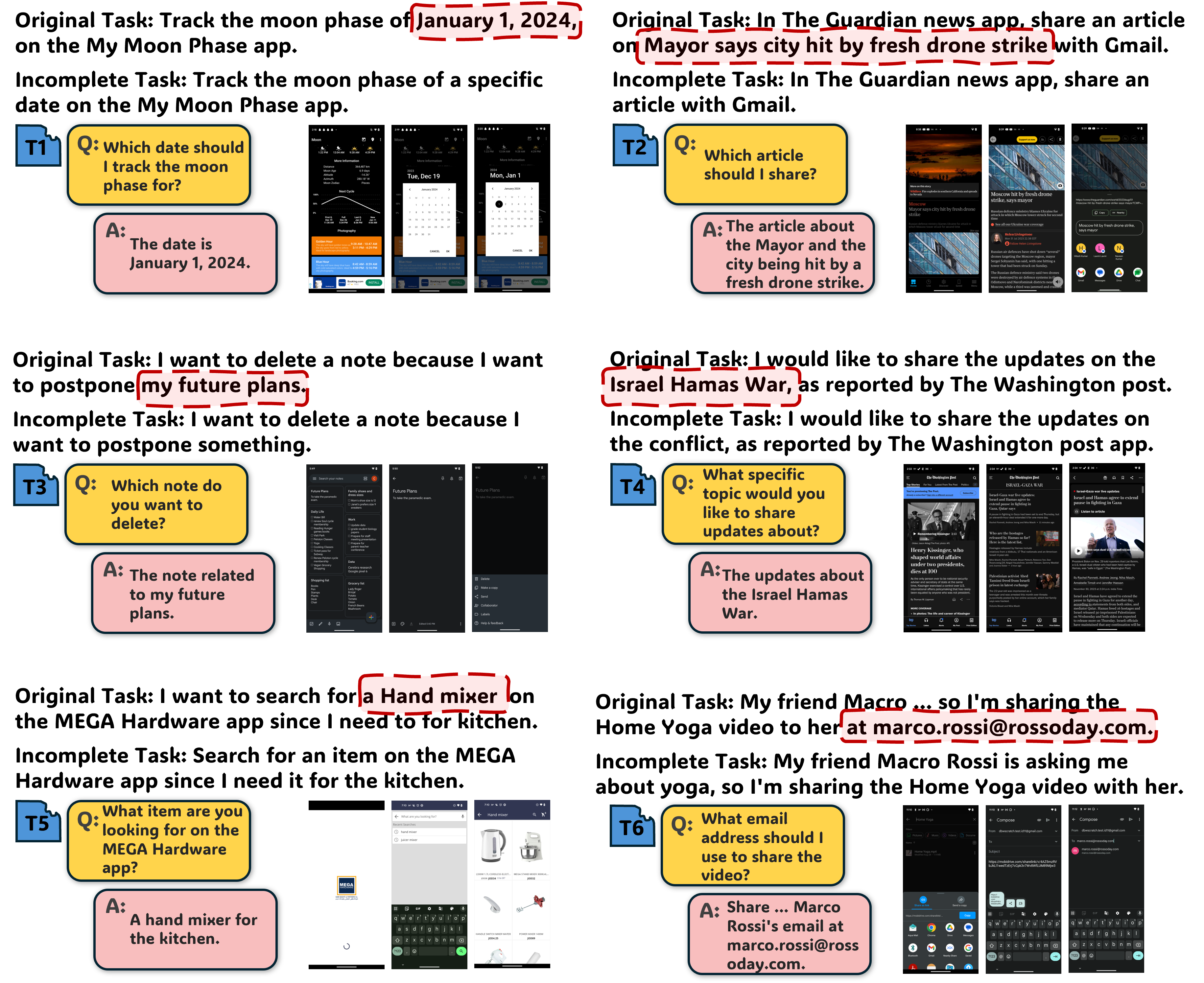}
  \caption{Demonstration of AndroidControl Navi\textit{plus} Dataset.}
  \label{fig:data-demo-androidcontrol}
\end{figure*}

%% file: latex/figures/data-demo-mind2web.tex
\begin{figure*}[ht]
  \centering
  \includegraphics[width=\linewidth]{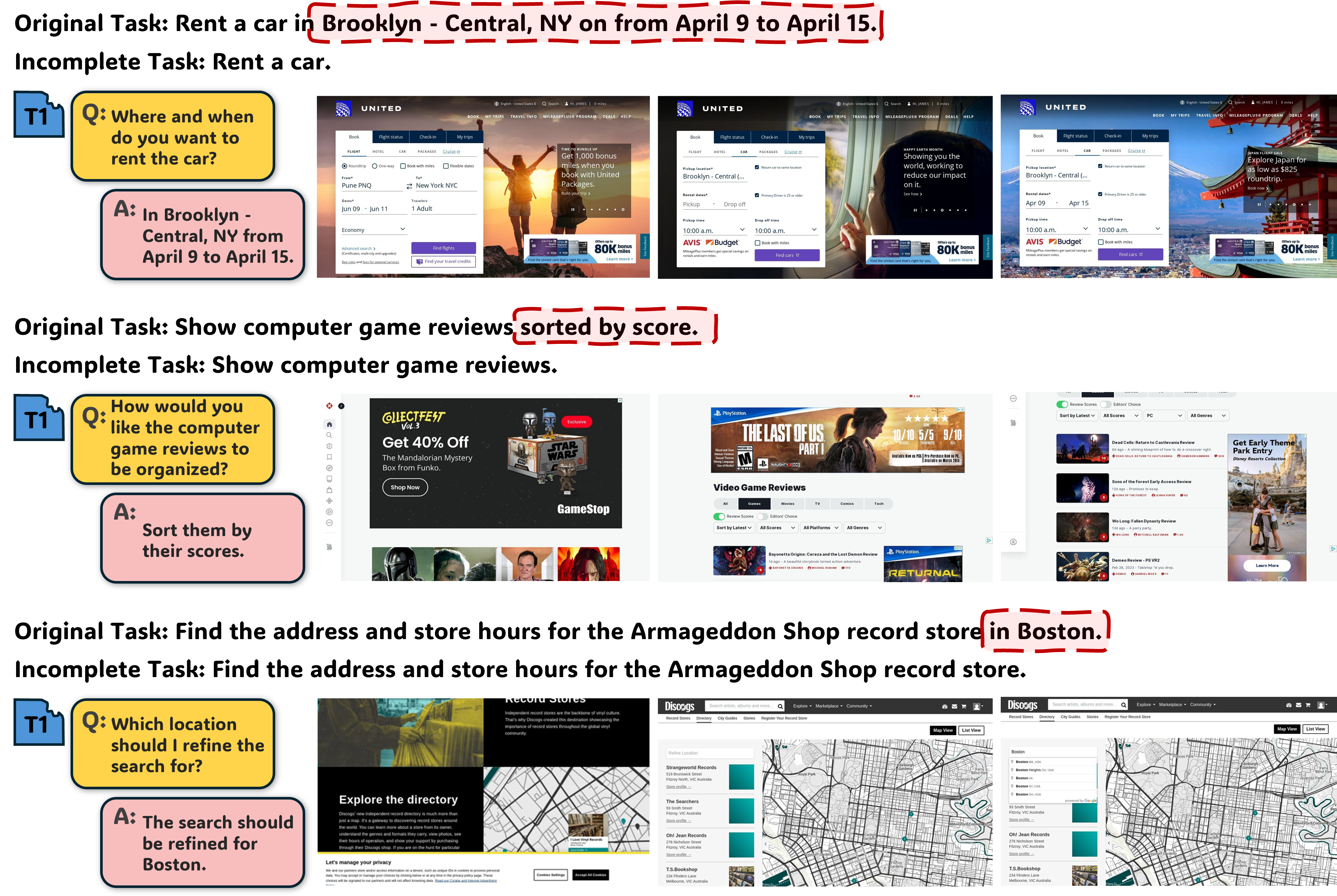}
  \caption{Demonstration of AndroidControl Navi\textit{plus} Dataset.}
  \label{fig:data-demo-mind2web}
\end{figure*}

%% file: latex/Appendix/_Appendix-Overview.tex
\appendix
\section*{Appendix Overview}

\noindent \textbf{Appendix \ref{appx:dataset-demo}:Examples of Navi\textit{plus} Dataset}

\noindent \textbf{Appendix \ref{appx:dataset-statistics}: Dataset Statistics}

\noindent \textbf{Appendix \ref{appx:extended-experiment-results}: Extended Experiment Results}

\noindent \textbf{Appendix \ref{appx:evaluation-metrics-formula}: Evaluation Metrics Formulas}

\noindent \textbf{Appendix \ref{appx:extended-related-work}: Extended Related Work}
  
\noindent \textbf{Appendix \ref{appx:prompt-templates}: Prompt Templates}

\noindent \textbf{Appendix \ref{appx:ethical-considerations}: Ethical Considerations}

%% file: latex/Appendix/A-Qualitative-Examples-of-Naviplus-Dataset.tex
\section{Qualitative Examples of Navi\textit{plus} Dataset}
\label{appx:dataset-demo}

Figure~\ref{fig:data-demo-androidcontrol} illustrates samples from the AndroidControl-Navi\textit{plus} dataset, and Figure~\ref{fig:data-demo-mind2web} illustrates samples from the Mind2Web-Navi\textit{plus} dataset.

%% file: latex/Appendix/B-Dataset-Statistics.tex
\section{Dataset Statistics}
\label{appx:dataset-statistics}


Figure~\ref{fig:data-distribution} shows some key statistics from our constructed Navi\textit{plus} dataset. 

As in the histogram of the lengths of the tasks (in steps) depicted in Figure~\ref{fig:sub:androidcontrol-steps} and Figure~\ref{fig:sub:mind2web-steps}, the distribution of the orange bars (representing the number of informative steps) is left-skewed compared to the blue bars (representing the number of all steps). On average for both AndroidControl-Navi\textit{plus} and Mind2Web-Navi\textit{plus} datasets, the informative steps are noticeably fewer than the steps for full tasks. This difference shows our data construction pipeline effectively distinct the informative and transforative steps.

When our data construction pipeline removes the informative steps from the full task, the informative steps of a task might be not enough for the given steps remove number. Figure~\ref{fig:sub:androidcontrol-enough} and Figure~\ref{fig:sub:mind2web-enough} displays the percentage of enoughed tasks and not-enoughed tasks in relation to the number of steps removed. Around Step Removed equals 2 to 4, the distribution shifts, where "Not Enough" becomes more prevalent. Within our Navi\textit{plus} dataset, the step removal is limited to two, as further deletion makes nearly half of the tasks overly general and meaningless.




%% file: latex/Appendix/C-Extended-Experiment-Results.tex
\section{Extended Experiment Results}
\label{appx:extended-experiment-results}


\subsection{Hyper-parameter analysis}

We conducted a hyperparameter analysis on SpiritSight-Agent-8B with LoRA ranks set to 16 and 64, training for epochs ranging from 0 to 2, and recorded steps from 200 to 2600 at intervals of 200. For a fair comparison with the baseline models from the original papers and the convenience of following works, we report the results of training 1 epoch. The results are presented in Figure~\ref{fig:hyperpara-androidcontrol}

\subsection{Extended Results on SpiritSight-Agent}

As shown in Table~\ref{scale-model-data-full-appx}, we provide the full experiment result for the scaling effect of model and dataset here in Section~\ref{sec:data-model-scale} and Section~\ref{sec:sr-results}.

\input{latex/tables/scale-model-data-full-appx}

%% file: latex/tables/scale-model-data-full-appx.tex
\begin{table*}[b]
\begin{center}
\small
\begin{tabular}{c
>{\columncolor[HTML]{EFEFEF}}c 
>{\columncolor[HTML]{EFEFEF}}c cccccccc}
\hline
                                       & \multicolumn{2}{c}{\cellcolor[HTML]{EFEFEF}w/ Original Task} & \multicolumn{8}{c}{w/ Incomplete Task +ASK +Dual Eval}                                                                                                                   \\ \cline{2-11} 
\multirow{-2}{*}{100\% Min2Web}        & SSR                           & SR                           & SSR                                                     & SR                                                       & Precision & Acc   & FPR   & F1    & CosSim & Meteor \\ \hline
SS-Agent-2B                            & 41.2                          & 6.8                          & \begin{tabular}[c]{@{}c@{}}39.5\\ (95.8\%)\end{tabular} & \begin{tabular}[c]{@{}c@{}}5.7\\ (83.8\%)\end{tabular}   & 0.833     & 0.896 & 0.009 & 0.438 & 0.547  & 0.360  \\
SS-Agent-8B                            & 47.2                          & 8.5                          & \begin{tabular}[c]{@{}c@{}}44.3\\ (93.9\%)\end{tabular} & \begin{tabular}[c]{@{}c@{}}5.9\\ (69.4\%)\end{tabular}   & 0.815     & 0.895 & 0.011 & 0.442 & 0.605  & 0.398  \\
SS-Agent-26B                           & 51.7                          & 9.9                          & \begin{tabular}[c]{@{}c@{}}47.7\\ (92.3\%)\end{tabular} & \begin{tabular}[c]{@{}c@{}}8.8\\ (88.8\%)\end{tabular}   & 0.810     & 0.894 & 0.011 & 0.435 & 0.622  & 0.447  \\ \hline\hline
                                       & \multicolumn{2}{c}{\cellcolor[HTML]{EFEFEF}w/ Original Task} & \multicolumn{8}{c}{w/ Incomplete Task +ASK +Dual Eval}                                                                                                                   \\ \cline{2-11} 
\multirow{-2}{*}{100\% AndroidControl} & SSR                           & SR                           & SSR                                                     & SR                                                       & Precision & Acc   & FPR   & F1    & CosSim & Meteor \\ \hline
SS-Agent-2B                            & 64.9                          & 20.5                         & \begin{tabular}[c]{@{}c@{}}63.4\\ (97.7\%)\end{tabular} & \begin{tabular}[c]{@{}c@{}}19.6\\ (95.6\%)\end{tabular}  & 0.917     & 0.909 & 0.007 & 0.588 & 0.778  & 0.689  \\
SS-Agent-8B                            & 67.6                          & 24.2                         & \begin{tabular}[c]{@{}c@{}}67.2\\ (99.4\%)\end{tabular} & \begin{tabular}[c]{@{}c@{}}23.1\\ (95.5\%)\end{tabular}  & 0.935     & 0.912 & 0.005 & 0.603 & 0.807  & 0.732  \\
SS-Agent-26B                           & 68.7                          & 24.0                         & \begin{tabular}[c]{@{}c@{}}67.6\\ (98.4\%)\end{tabular} & \begin{tabular}[c]{@{}c@{}}23.8\\ (99.2\%)\end{tabular}  & 0.934     & 0.911 & 0.005 & 0.593 & 0.801  & 0.719  \\ \hline\hline
                                       & \multicolumn{2}{c}{\cellcolor[HTML]{EFEFEF}w/ Original Task} & \multicolumn{8}{c}{w/ Incomplete Task +ASK +Dual Eval}                                                                                                                   \\ \cline{2-11} 
\multirow{-2}{*}{10\% AndroidControl}  & SSR                           & SR                           & SSR                                                     & SR                                                       & Precision & Acc   & FPR   & F1    & CosSim & Meteor \\ \hline
SS-Agent-2B                            & 52.8                          & 11.0                         & \begin{tabular}[c]{@{}c@{}}50.7\\ (96.0\%)\end{tabular} & \begin{tabular}[c]{@{}c@{}}9.3\\ (84.5\%)\end{tabular}   & 0.823     & 0.890 & 0.013 & 0.474 & 0.700  & 0.596  \\
SS-Agent-8B                            & 59.9                          & 15.8                         & \begin{tabular}[c]{@{}c@{}}57.4\\ (95.8\%)\end{tabular} & \begin{tabular}[c]{@{}c@{}}12.6\\ (79.7\%)\end{tabular}  & 0.881     & 0.887 & 0.007 & 0.429 & 0.715  & 0.608  \\
SS-Agent-26B                           & 60.2                          & 15.9                         & \begin{tabular}[c]{@{}c@{}}58.4\\ (97.0\%)\end{tabular} & \begin{tabular}[c]{@{}c@{}}15.9\\ (100.0\%)\end{tabular} & 0.794     & 0.874 & 0.010 & 0.336 & 0.707  & 0.607  \\ \hline
\end{tabular}
\end{center}
\caption{Performance of the SpiritSight-Agent in terms of SSR and SR across model sizes (2B, 8B, and 26B) and different task configurations.}
\label{scale-model-data-full-appx}
\end{table*}

%% file: latex/Appendix/D-Evaluation-Metrics-Formulas.tex
\section{Evaluation Metrics Formula}
\label{appx:evaluation-metrics-formula}

\subsection{SSR (Step Success Rate)}

For each episode \( e \), we calculate the ratio of correct steps \( C(e) \) to the total number of steps \( T_e \). The Step Success Rate (SSR) is then computed as the average of this ratio over all episodes:

\[
\text{SSR} = \frac{1}{N} \sum_{e=1}^{N} \frac{C(e)}{T_e}
\]

Where:
\( N \) is the total number of episodes.
\( C(e) \) is the number of correct steps in episode \( e \).
\( T_e \) is the total number of steps in episode \( e \).

\subsection{SSR before/after}

For each episode \( e \), we calculate the Step Success Rate (SSR) for the steps before and after a fixed ASK step at position \( k \). The SSR before and after are then calculated as the averages across all episodes.

\[
\text{SSR}_{\text{before}} = \frac{1}{N} \sum_{e=1}^{N} \frac{C_{\text{before}}(e)}{T_{\text{before}}(e)}
\]

Where:
\( C_{\text{before}}(e) \) is the number of correct steps before the ASK step in episode \( e \).
\( T_{\text{before}}(e) \) is the total number of steps before the ASK step in episode \( e \).

Similarity for the steps after the ASK step:

\[
\text{SSR}_{\text{after}} = \frac{1}{N} \sum_{e=1}^{N} \frac{C_{\text{after}}(e)}{T_{\text{after}}(e)}
\]

Where:
\( C_{\text{after}}(e) \) is the number of correct steps after the ASK step in episode \( e \).
\( T_{\text{after}}(e) \) is the total number of steps after the ASK step in episode \( e \).

\subsection{SR (Success Rate)}

For each episode \( e \), we check whether each step \( k \) is correct. If all steps in an episode are correct, the episode is considered successful. The Success Rate (SR) is then computed as the ratio of successful episodes to the total number of episodes:

\[
\text{SR} = \frac{1}{N} \sum_{e=1}^{N} \mathbb{I}_{\text{success}}(e)
\]

Where:
\( N \) is the total number of episodes.
\( \mathbb{I}_{\text{success}}(e) \) is the indicator function, where \( \mathbb{I}_{\text{success}}(e) = 1 \) if episode \( e \) is successful (all steps correct), and \( \mathbb{I}_{\text{success}}(e) = 0 \) otherwise.

\subsection{CosSim (Cosine Similarity)}

Given two ASK sentences \( s_1 \) and \( s_2 \), we compute their embeddings \( \mathbf{e}_1 \) and \( \mathbf{e}_2 \) using a Sentence Transformer model:

\begin{align}
\mathbf{e}_1 &= \text{SentenceTransformer}(s_1) \\
\mathbf{e}_2 &= \text{SentenceTransformer}(s_2)
\end{align}

The cosine similarity between the two embeddings is then calculated as:

\[
\text{CosSim}(\mathbf{e}_1, \mathbf{e}_2) = \frac{\mathbf{e}_1 \cdot \mathbf{e}_2}{\|\mathbf{e}_1\| \|\mathbf{e}_2\|}
\]

Where:
\( \mathbf{e}_1 \cdot \mathbf{e}_2 \) is the dot product of the two embeddings.
\( \|\mathbf{e}_1\| \) and \( \|\mathbf{e}_2\| \) are the magnitudes of the embeddings.

%% file: latex/Appendix/E-Extanded-Related-Work.tex
\section{Extended Related Work}
\label{appx:extended-related-work}

\subsection{GUI Navigation Agents and Datasets}
Research on GUI navigation agents is gaining popularity, and the resulting works are becoming more diverse and powerful. The recent introduction of Vision Language Model (VLM) based methods, such as SeeClick \citep{cheng2024seeclick} and SeeAct \citep{zheng2024seeact}, has revolutionized the original text-only Large Language Models (LLMs) methods \citep{deng2024mind2web} by providing more compact and informative screenshots as primary context sources.

Abundant data resources have been gathered to motivate the comprehensive development of VLM-powered GUI agents. These efforts have generally three levels: (1) GUI environment understanding, (2) Operational elements grounding, and (3) Navigation task planning.

\textbf{(1) GUI environment understanding}

Works such as RICOSCA\citep{deka2017rico}, Widget Captioning\citep{li2020Widget-captioning}, Screen2Words\citep{wang2021screen2words}, ScreenQA\citep{hsiao2022screenqa}, WebSRC\citep{chen2021websrc}, GUICourse\citep{chen2024guicourse}, MP-GUI\citep{wang2025mpgui} have constructed screenshot-conditioned question-answering data to enhance the agents' domain knowledge in GUI environments.

\textbf{(2) Operational elements grounding}

Some works make great efforts in collecting large-scale screenshots annotated with element content and locations to enhance agents' GUI element grounding capability. A variety of GUI platforms have been comprehensively covered: datasets like SeeClick\citep{cheng2024seeclick}, GUICourse\citep{chen2024guicourse}, AguVis\citep{xu2024aguvis},, SpiritSight\citep{huang2025spiritsight} focuses on web pages. AMEX\citep{chai2024amex}, FerretUI-v1\citep{you2024ferret-ui-v1}, FerretUI-v2\citep{li2024ferret-ui-v2} incorporates extensive mobile devices, OS-ATLAS\citep{wu2024os-atlas}, ShowUI\citep{lin2024showui}, DeskVision\citep{xu2025deskvision} collected in complicated personal computer desktops.

Furthermore, AMEX\citep{chai2024amex}, GUICourse\citep{chen2024guicourse}, SpiritSight\citep{huang2025spiritsight}, MP-GUI\citep{wang2025mpgui}, ShowUI\citep{lin2024showui} have annotated the functionality of GUI elements to enable smoother generalization from element-level tasks to navigation tasks.

\textbf{(3) Navigation task planning}

Studies like Mind2Web\citep{deng2024mind2web}, OmniAct\citep{kapoor2024omniact}, GUICourse\citep{chen2024guicourse}, MoTIF\citep{burns2022motif}, AITW\citep{rawles2024AndroidintheWild}, GUI-Odyssey\citep{lu2024GUI-Odyssey}, AMEX\citep{chai2024amex}, AndroidControl\citep{li2024androidcontrol}, OS-Genesis\citep{sun2024os-genesis} have annotated real GUI navigation trajectories on various platforms for training and benchmarking practical GUI automation agents. 

AITZ\citep{zhang2024AndroidintheZoo}, AutoGLM\citep{liu2024autoglm}, CogAgent-V2\citep{hong2024cogagent}, InfiGUIAgent\citep{liu2025infiguiagent} have augmented agents' task planning capabilities by using CoAT approaches and self-reflections to empower scaling at inference-time. UITARS\citep{qin2025uitars} and UI-R1\citep{lu2025uir1} further applied reinforce learning techniques like DPO and GRPO to directly optimize agents' reasoning on out-of-distribution OOD tasks.

%% file: latex/Appendix/F-Prompt-Templates.tex
\section{Prompt Templates}
\label{appx:prompt-templates}

\subsection{Low-level Instruction Completion}
For prompt template for low-level instruction completion step, see Figure~\ref{prompt-low-ins}.

\begin{figure*}[htbp]
    \centering
    \begin{tcolorbox}[colframe=blue!20, colback=blue!10, coltitle=black, title=Prompt Template for Low-level Instruction Completion]
    \textbf{Prompt:}\\
    <image> \\
    I will provide you with a full task description that is completed by performing a series of actions within web browser. These actions are performed sequentially as steps, and together they result in an operation trajectory for completing the task. Your mission is to generate a step instruction with the given a action code and the current screenshot content. \\
    
    The action code indicates the clicked content or inputted content. The generated Step Instruction should be concise, directly related to the action code and its purpose. If the action code is CLICK(UnKnown), you need to identify the content in the red bbox to know what is exactly clicked.\\
    
    Please output in JSON format, structured as follows:\\
    \{\\
        "Full Task": "<Complete task description here>",\\
        "Step Action Code": "<the provided action code>",\\
        "Red Bbox Content": "<the content inside the red bbox>",\\
        "Step Instruction": "<Generate a Step Instruction here>",\\
    \}\\
    Do not output other explanations.\\
    
    \#\# Full Task: \{task\}\\
    \#\# Step Action Code: \{action\}
    \end{tcolorbox}
    \caption{Prompt Template for Low-level Instruction Completion.}
    \label{prompt-low-ins}
\end{figure*}

\subsection{Informative Step Decision \& Formation of Ambiguous Tasks}
For prompt template for informative step decision and formation of ambiguous tasks step, see Figure~\ref{prompt-make-task}.

\begin{figure*}[htbp]
    \centering
    \begin{tcolorbox}[colframe=blue!20, colback=blue!10, coltitle=black, title=Prompt Template for Informative Step Decision \& Formation of Ambiguous Tasks]
    \textbf{Prompt:}\\
    You are a task simplification Specialist. You should maintain outputting accurate sentences. 
    I will provide you with a full task description and one step instruction in this trajectory.
    The full task description is a GUI operation task completed by sequentially performing a series of steps within mobile phone apps.\\
    
    Your mission is to create a **Simplified Task Description** by removing the information contained in a selected step instruction from the full task description.\\
    
    \#\#\# Please follow this step by step solution: \\
    1. Repeat the full task description and the selected step instruction to make sure you understand the input information.\\
    2. Find the overlapping information of the selected step instruction within the full task, base on named entity with specific information.\\
    3. Form the simplified task description by removing the overlapping information from the full task description while making sure the remaining content unchanged. Only remove the related words, or replace specific entity with reference word.\\
    4. Rephrase the generated simplified task description using the same imperative tone and style as the original task if necessary.\\
    5. Generate a follow-up question to simulate as if the agents tries to clarify about the removed information.\\
    6. Generate the human's clarifying answer based on the step instruction removed, do not straightly say the operation, but only say about the intention.\\
    
    Please ensure that the output is in JSON format, structured as follows:\\
    \{\\
        "Full Task": "<Complete task description here>",\\
        "Selected Step to Exclude": "<Step to be removed here>",\\
        "Overlapping Information": "<Details of the task description that overlap with the selected step. If no specific information is overlapped, say 'None'>",\\
        "Incomplete Task Description": "<Generated task description without the selected step's information>",\\
        "Rephrased Incomplete Task Description": "<If any rephrasing is required, show the final version here>",\\
        "Follow Up Question": "<Generate a follow-up question that asks about the removed step>",\\
        "Human Answer": "<Generate the human's clarifying answer, do not include operation>"\\
    \}\\
    
    \#\# Examples: \{Few-shot Examples\}
    \end{tcolorbox}
    \caption{Prompt Template for Informative Step Decision \& Formation of Ambiguous Tasks}
    \label{prompt-make-task}
\end{figure*}

%% file: latex/Appendix/G-Ethical-Considerations.tex
\section{Ethical Considerations}
\label{appx:ethical-considerations}

GUI automation agents have significant social, security, and privacy implications. On the one hand, they can free humans from repetitive operational tasks on digital devices and enhance work efficiency. On the other hand, if they fall into malicious hands, GUI agents could be misused to bypass anti-fraud systems or manipulate software to achieve harmful or unintended results.   Additionally, GUI agents may make errors while performing tasks, leading to unacceptable results or unwanted side effects. There is also the risk of leaked private information if proper data collection regulations are not in place. For these reasons, GUI automation agents must be fully regulated to ensure that their broader use serves the social good.